\documentclass{article}

\usepackage{microtype}
\usepackage{graphicx}
\usepackage{subfigure}

\usepackage{booktabs} %

\usepackage{hyperref}
\usepackage{paralist}

\makeatletter
\def\url@leostyle{%
  \@ifundefined{selectfont}{\def\UrlFont{}}%
  {\def\UrlFont{}}%
}
\makeatother
\usepackage[accepted]{icml2022}

\usepackage{amsmath}
\usepackage{amssymb}
\usepackage{mathtools}
\usepackage{amsthm}

\usepackage[capitalize,noabbrev]{cleveref}

\theoremstyle{plain}

\theoremstyle{definition}

\theoremstyle{remark}

\usepackage[textsize=tiny]{todonotes}

\usepackage[flushmargin]{footmisc}
\usepackage{bbold}
\usepackage{color}
\usepackage[small,bf]{caption}

\usepackage{graphicx} %
	\setkeys{Gin}{width=\textwidth, height=\textheight, keepaspectratio}
\newif\ifcomment
\commenttrue

\ifcomment
	\newcommand{\edc}[1]{\textbf{\em\color{red}EDC: #1}}
	\newcommand{\bo}[1]{\textbf{\em\color{green}BO: #1}}
	\newcommand{\gvg}[1]{\textbf{\em\color{blue}GG: #1}}
\else
		\newcommand\edc[1]{}
		\newcommand\bo[1]{}
		\newcommand\gvg[1]{}
\fi
\newcommand{\descr}[1]{\smallskip\noindent\textbf{#1}}
\newcommand{\descrit}[1]{\vspace{0.001cm}\noindent\textit{#1}}

 \definecolor{linkcol}{RGB}{0,70,25}
 \definecolor{citecol}{RGB}{0,70,25}
 \definecolor{urlcol}{RGB}{70,0,25}
\icmltitlerunning{Differential Privacy Has Disparate Impact}

\begin{document}

\twocolumn[
\icmltitle{\bf {\em Robin Hood} and {\em Matthew} Effects: Differential Privacy Has\\ Disparate Impact on Synthetic Data}

\icmlsetsymbol{equal}{*}

\begin{icmlauthorlist}
\icmlauthor{Georgi Ganev}{ucl,hazy}
\icmlauthor{Bristena Oprisanu}{ucl}
\icmlauthor{Emiliano De Cristofaro}{ucl}
\end{icmlauthorlist}

\icmlaffiliation{ucl}{University College London, London, UK}
\icmlaffiliation{hazy}{Hazy, London, UK}

\icmlcorrespondingauthor{Georgi Ganev}{georgi.ganev.16@ucl.ac.uk}

\vskip 0.2in
]

\printAffiliationsAndNotice{} %

\begin{abstract}
Generative models trained with Differential Privacy (DP) can be used to generate synthetic data while minimizing privacy risks.
We analyze the impact of DP on these models vis-\`a-vis {\em underrepresented} classes/subgroups of data, specifically, studying: 1) the {\em size} of classes/subgroups in the synthetic data, and 2) the {\em accuracy} of classification tasks run on them.
We also evaluate the effect of various levels of imbalance and privacy budgets.
Our analysis uses three state-of-the-art DP models (PrivBayes, DP-WGAN, and PATE-GAN) and shows that DP yields opposite size distributions in the generated synthetic data.
It affects the gap between the majority and minority classes/subgroups; in some cases by reducing it (a ``Robin Hood'' effect) and, in others, by increasing it (a ``Matthew'' effect).
Either way, this leads to (similar) disparate impacts on the accuracy of classification tasks on the synthetic data, affecting disproportionately more the underrepresented subparts of the data.
Consequently, when training models on synthetic data, one might incur the risk of treating different subpopulations unevenly, leading to unreliable or unfair conclusions.
\end{abstract}

\section{Introduction}
\label{sec:intro}

Releasing synthetic data is an increasingly advocated and adopted approach to reduce privacy risks while sharing data~\cite{schaar20synthetic}.
Synthetic data initiatives %
have been promoted by, e.g., the US Census Bureau~\cite{benedetto2018creation}, England's National Health Service~\cite{nhs21ae}, and NIST%
~\cite{nist2018differential, nist2018the}.

The idea is to train {\em generative} machine learning models to learn the probabilistic distribution of the (real) data and then sample from the model to generate new (synthetic) data records.
However, real-world datasets often contain personal and sensitive information about individuals~\cite{thompson19twelve} that could leak into/through models that are trained on them.
Generative models can overfit or memorize individual data points~\cite{carlini2019secret, webster2019detecting}, %
which facilitates privacy attacks such as membership or property inference attacks~\cite{hayes2019logan, chen2020gan, stadler2022synthetic}.

The state-of-the-art method for training models that provably minimize inferences is to do while satisfying Differential Privacy (DP)~\cite{dwork2014algorithmic}.
DP provides a mathematical guarantee on the privacy of all records in the training dataset by bounding their individual contribution.
This can be achieved by applying noise (e.g., using the Laplace Mechanism~\cite{dwork2006calibrating}), relying on techniques such as DP-Stochastic Gradient Descent (DP-SGD)~\cite{abadi2016deep}, or Private Aggregation of Teacher Ensembles (PATE)~\cite{papernot2016semi, papernot2018scalable}.

Naturally, as they rely on perturbation, DP methods inherently reduce accuracy in the task the data is used for.
Incidentally, this degradation is often disproportionate; for instance, the accuracy of {\em DP classifiers} often drops more for the underrepresented classes and subgroups of the dataset.
Prior work~\cite{bagdasaryan2019differential, farrand2020neither,uniyal2021dp} illustrates this effect when deep neural networks are trained with DP-SGD or PATE on imbalanced datasets.
Moreover, {\em DP statistics} have also been shown to lead to disproportionate biases~\cite{kuppam2019fair}.

\descr{Problem Statement.}
So far, this ``disparate effect'' caused by DP and its applications have only been analyzed in the context of discriminative models.
This paper focuses on DP generative models and tabular synthetic data. %
We look at the problem from two angles: 1) {\em counts comparisons} and 2) {\em downstream tasks} such as classification.
We analyze three widely used DP generative models: PrivBayes~\cite{zhang2017privbayes}, DP-WGAN~\cite{alzantot2019differential}, and PATE-GAN~\cite{jordon2018pate}, which rely, respectively, on the Laplace Mechanism, DP-SGD, and PATE. %

Our work aims to answer the following research questions:
\begin{compactitem}
	\item[--] {\bf RQ1:} Do DP generative models generate data in similar classes and subgroups proportions to the real data?\smallskip
	\item[--] {\bf RQ2:} Does training a classifier on DP synthetic data lead to the same disparate impact on accuracy as training a DP classifier on the real data?\smallskip
	\item[--] {\bf RQ3:} Do different DP mechanisms for DP synthetic data behave similarly under different privacy and data imbalance levels?
\end{compactitem}

\descr{Main Findings.}
Overall, our experiments show that:%
\begin{compactenum}
	\item There is a disparate effect on the classes and subgroups sizes in the synthetic data generated by all DP generative models.
	This effect is dependent on the generative model and DP mechanism; e.g., PrivBayes evens the data, while PATE-GAN increases the imbalance.\smallskip
	\item There also is a disparate effect on the accuracy of classifiers trained on synthetic data generated by all generative models; for instance, underrepresented classes and subgroups suffer bigger and/or more variable drops.
	Furthermore, majority classes with similar characteristics to minority classes could also suffer from a disproportionate drop in utility.
	\smallskip
	\item The magnitude of these effects on size and accuracy increases when stronger privacy guarantees are imposed.
	Higher data imbalance levels further intensify them.
	Also, some generative models are better suited for specific privacy budgets and imbalance levels.\smallskip
	\item While classifiers trained on data generated by PATE-GAN perform much better than, or on par with, DP-WGAN, we observe some undesirable behaviors: PATE-GAN completely fails to learn some subparts of the data with highly imbalanced multi-class data. With low privacy budgets, it also generates synthetic data with artificially enhanced correlation between the subgroup and the target columns.
\end{compactenum}

\section{Preliminaries}
In this section, we present some background information, %
then, we introduce the datasets, the classifiers used for baselines, and the generative models used for producing synthetic data. %
(The source of the implementations we use, when applicable, is also reported.)

\subsection{Background}
\label{sec:background}

\noindent{\bf Generative Models and Synthetic Data.}
During fitting, the generative model training algorithm $GM(D^n)$ takes in input $D^n$ (a sample dataset consisting of $n$ records drawn iid from the population $D^n{\sim}P(\mathbb{D})$), updates its internal parameters to learn $P_g(D^n)$, a (lower-dimensional) representation of the joint probability distribution of the sample dataset $P(D^n)$, and outputs a trained model $g(D^n)$.
Then, one can sample from the trained model to generate a synthetic dataset of size $m$, $S^m{\sim}P(g(D^n))$.
Both the fitting and generation steps are stochastic; in order to get confidence intervals, one can train the generative model $l$ times and sample $k$ synthetic datasets for each trained model.

While several different approaches exist to build generative models, in this paper, we focus on two of them, specifically: 1) Bayesian networks~\cite{koller2009probabilistic, barber2012bayesian}, and 2) Generative Adversarial Networks (GANs)~\cite{goodfellow2014generative}.
The former is a graphical model that breaks down the joint distribution by explicit lower-dimensional conditional distributions.
The latter approximates the dataset distribution implicitly by iteratively optimizing a min-max ``game'' between two neural networks: a generator, producing synthetic data, and a discriminator, trying to distinguish real from synthetic samples.

\descr{Differential Privacy (DP).}
Let $\epsilon$ be a positive and real number and $\mathcal{A}$ a randomized algorithm.
$\mathcal{A}$ satisfies $\epsilon$-DP if, for all neighboring datasets $D_{1}$ and $D_{2}$ (differing in a single data record), and all possible outputs $S$ of $\mathcal{A}$, the following holds~\cite{dwork2014algorithmic}:
\begin{equation*}
P[{\mathcal{A}}(D_{1})\in S]\leq \exp \left(\epsilon \right)\cdot P[{\mathcal{A}}(D_{2})\in S]
\end{equation*}
In other words, looking at the output of the algorithm, one cannot distinguish whether any individual's data was included in the input dataset or not.
The level of that indistinguishability is measured by $\epsilon$, also called a privacy budget.

In the context of machine learning, $\mathcal{A}$ is usually the training procedure. %
In this paper, we focus on three DP techniques: the Laplace mechanism~\cite{dwork2006calibrating}, DP-SGD~\cite{abadi2016deep}, and PATE~\cite{papernot2016semi, papernot2018scalable} (for more details, see Sec.~\ref{ssec:GMs}).
The last two techniques use a relaxation of DP called ($\epsilon$, $\delta$)-DP~\cite{dwork2014algorithmic}; here, $\delta$, usually a small number, denotes a probability of failure.
Finally, due to its robustness to post-processing, DP allows for DP-trained models to be re-used without further privacy leakage.

\descr{Disparate Impact Metrics.}
For the downstream task evaluation (see Sec.~\ref{ssec:EM}), we follow the disparate impact metrics proposed in~\cite{bagdasaryan2019differential} and use {\em accuracy parity}, a weaker form of ``equal odds''~\cite{hardt2016equality}.
Specifically, we focus on model accuracy on imbalanced classes (and multi-classes) and imbalanced subgroups (with balanced classes) of the dataset.
Similarly to~\cite{bagdasaryan2019differential}, we do not consider (other) fairness evaluations, leaving them as items for future work. %

\subsection{Datasets}
We consider several tabular and one image datasets from different domains, which are widely used in the ML research community.
All have an associated classification task or have slightly been modified for this purpose.

\descr{Adult.}
The Adult dataset~\cite{dua2017adult} is extracted from the 1994 Census database, consisting of 32,561 training and 16,281 testing records.
It has 15 attributes: 6 numerical, including age, and 9 categorical, including sex and race.
The target column indicates whether the individual's income exceeds \$50K/year.

\descr{Texas.}
The Texas Hospital Inpatient Discharge dataset~\cite{dshs2013texas} contains data on discharges from Texas hospitals.
As done in previous work~\cite{stadler2022synthetic}, we sample 49,983 records from 2013 and select 12 attributes, 1 numerical and 11 categorical, including age, sex, and race.
To create a classification task, we convert the numerical attribute, indicating the length of stay in the hospital, into a categorical one by specifying whether the person's hospitalization was a week or longer.

\descr{Purchases.}
The Purchases dataset is based on Kaggle's ``Acquire Valued Shoppers Challenge''~\cite{kaggle2013purchases}, aimed at predicting whether customers would become loyal to products based on incentives.
As done in previous work~\cite{shokri2017membership}, we modify the main task to be predicting customers' purchase style.
First, we filter products purchased at least 750,000 times and customers who made at least 500 purchases.
Then, we summarize the data so that each row represents a customer with 108 binary features, corresponding to whether the customer has bought that product or not.
Finally, to create the different purchasing styles, we cluster the customers into 20 clusters using a Mixture of Gaussian models.
This yields a dataset with 152,369 customers and 109 attributes, including the style.
Unlike the previous datasets, the classification task here is multi-class.

\descr{MNIST.}
The MNIST dataset~\cite{lecun2010mnist} consist of 60,000 training and 10,000 testing black and white handwritten 784-pixel digits.
The goal is to classify the digit, making it a multi-class problem with 10 classes.

\subsection{Generative Models}
\label{ssec:GMs}

Our evaluation includes three of the most popular DP generative models: a statistical one based on Bayesian networks and two GANs incorporating DP mechanisms.
Unless stated otherwise, we use the default hyperparameters, as provided by the authors.

\descr{PrivBayes.} %
PrivBayes~\cite{zhang2017privbayes, ping2017data} first constructs an optimal Bayesian network that approximates the joint data distribution by low-dimensional conditional distributions and then estimates them.
Both of these steps are done with $\epsilon$-DP guarantees, respectively, using the Exponential Mechanism~\cite{mcsherry2007mechanism} to choose the parents for each child node and the Laplace Mechanism~\cite{dwork2006calibrating} to construct noisy contingency tables before converting them to distributions.
Looking at the step involving Laplace Mechanism in more detail, any negative noisy counts are clipped at 0 before being normalized to a distribution, potentially leading to a biased estimator.
We discretize numerical columns to 50 bins, as opposed to 20, and set the degree of the network to 3 for all datasets except for Purchases, where it is 2.
Furthermore, we identified an industry-wide bug in the open-source package violating the DP guarantees and fixed it.\footnote{\url{https://github.com/DataResponsibly/DataSynthesizer/issues/34}}

\descr{DP-WGAN.} %
DP-WGAN~\cite{alzantot2019differential} is one of the top 5 solutions to the 2018 NIST Contest~\cite{nist2018differential}.
It relies on the WGAN architecture~\cite{arjovsky2017wasserstein}, which improves training stability and performance by using the Wasserstein distance instead of the Jensen-Shannon divergence as in GANs.
Furthermore, ($\epsilon$, $\delta$)-DP of the output is achieved using DP-SGD~\cite{abadi2016deep}, which sanitizes the gradients (clips the $\ell_2$ norm of the individual gradients and applies Gaussian Mechanism~\cite{dwork2006our} to the sum) of the discriminator during training.
The privacy budget is tracked using the moments accountant method~\cite{abadi2016deep}.
To be consistent with PATE-GAN (see below), we set $\delta=10^{-5}$ for all experiments.

\descr{PATE-GAN.} %
PATE-GAN~\cite{jordon2018pate} adapts the PATE framework~\cite{papernot2016semi, papernot2018scalable} for training GANs.
Instead of a single discriminator, there are $k$ teacher-discriminators and a student-discriminator.
The teacher-discriminators only see a disjoint partition of the real data. They are trained to minimize the classification loss when classifying samples as real or fake. In contrast, the student-discriminator is trained using noisy labels (using the Laplace Mechanism) predicted by the teachers.
As before, the privacy budget of the algorithm is calculated using the moments accountant~\cite{abadi2016deep} and the output is ($\epsilon$, $\delta$)-DP, with $\delta=10^{-5}$.

\subsection{Discriminative Models}
As mentioned, our experiments include a downstream task (classification) run on the synthetic data.
We use Logistic Regression (LR) to avoid another layer of stochasticity; specifically two versions of LR: the standard one in Scikit-Learn
~\cite{sklearn2011} and one with DP guarantees~\cite{chaudhuri2011differentially, holohan2019diffprivlib}.
The latter achieves $\epsilon$-DP by perturbing the objective function before optimization.

\section{Experimental Evaluation}

\subsection{Evaluation Methodology}
\label{ssec:EM}

Broadly, our goal is to empirically measure the impact of generative models with different DP mechanisms, $\epsilon$ levels, and data imbalance ratios have on class/subgroups distributions in the generated synthetic data and downstream task performance.
We consider four settings:
\begin{compactenum}
	\item[S1)] {\em Binary class size, precision, and recall.} We focus on the effect on binary classes,
	reporting class recall and precision because the target columns in all datasets are imbalanced.\smallskip
	\item[S2)] {\em Multi-class size, precision, and recall.} We study the effect on multi-classes.
	As in the previous setting, we report class recall and precision.\smallskip
	\item[S3)] {\em Single-attribute subgroup size, accuracy, and correlation.} We analyze the effect on a single-attribute subgroup.
	Here, we treat a single feature (e.g., sex) as a subgroup.
	We imbalance the dataset, so the minority subgroup comprises the desired ratio of the population while keeping the class per subgroup balanced.\smallskip
	\item[S4)] {\em Multi-attribute subgroup size, accuracy, and correlation.} We focus on the effect on multi-attribute subgroups.
	We treat an intersection of features (e.g., age, sex, and race) as small fine-grained subgroups.
	As in the previous setting, we balance the data only according to a single-attribute subgroup; otherwise, we risk throwing too much data out.
	Thus, we discard subgroups with fewer than 25 members.
\end{compactenum}

\smallskip
All evaluation settings follow three steps: dataset preparation, synthetic data generation, and prediction -- see below.

\descr{Dataset Preparation.}
First, we split the dataset into training and testing if the latter is not explicitly provided.
If the subgroup imbalance level is provided (S3 and S4), for both training and testing datasets, we balance the subgroup by class, so there are 50\% of each class per subgroup (we only consider binary classes in these settings).

Then, we imbalance the datasets to the desired subgroup imbalance level (while maintaining class parity), where the level represents the ratio of minority subgroup to the total size of the dataset.

\descr{Synthetic Data Generation.}
For a given generative model and privacy budget $\epsilon$, we train $l$ (we set $l=10$) generators and generate $k$ (we set $k=10$) synthetic datasets with size equal to the input dataset.
This results in $l\cdot{k}$ synthetic datasets.
We measure the class/subgroups distributions.
If single/multi-attribute subgroup is provided (S3 and S4), we also measure correlation between the subgroup and target columns by calculating the mutual information.

\descr{Classifiers Prediction.}
We capture the performance of three types of classifiers: 1) \emph{real classifier} -- we train a single LR on the real dataset and predict on the test dataset to serve as an overall baseline; 2) \emph{DP classifiers} -- we train $l\cdot{k}$ (equal to the number of synthetic datasets) DP LRs on the real dataset and predict on the test dataset; 3) \emph{synth classifiers} -- we train a single LR per synthetic dataset (in total $l\cdot{k}$) and predict on the test dataset.

\begin{figure*}[ht!]
	\centering
	\subfigure[\scriptsize PrivBayes]{\includegraphics[width=0.34\textwidth]{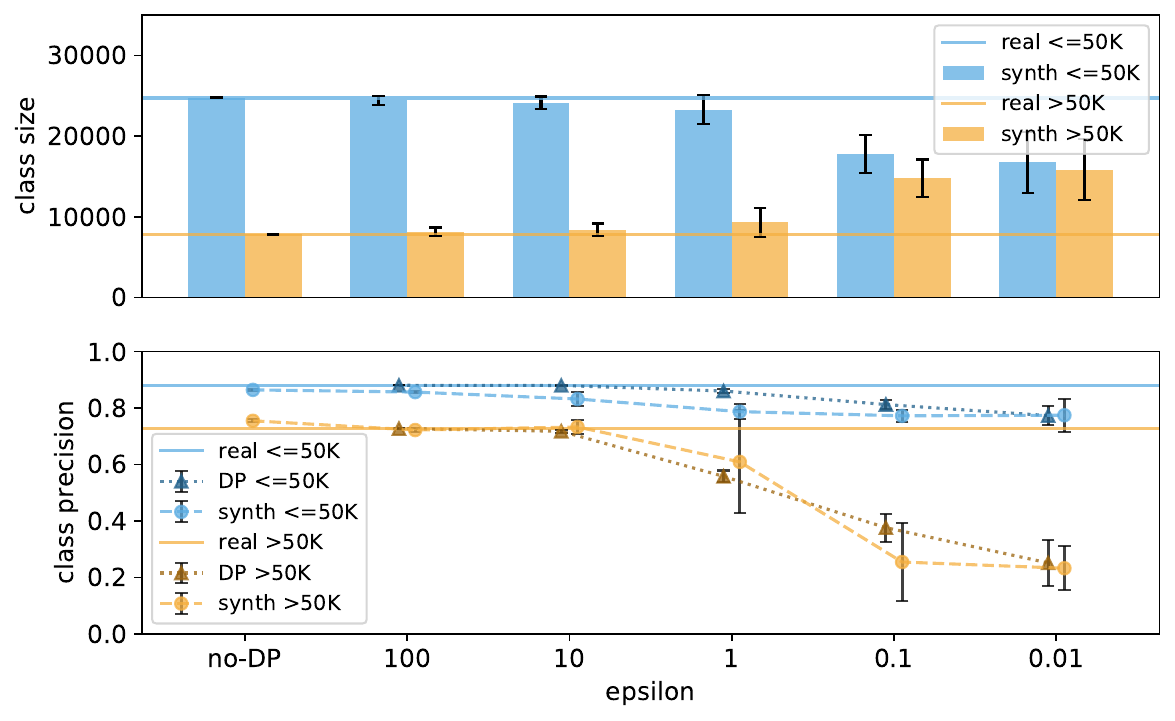}\label{fig:CSPR_A_PrivBayes}}\hspace*{-0.175cm}
	\subfigure[\scriptsize DP-WGAN]{\includegraphics[width=0.34\textwidth]{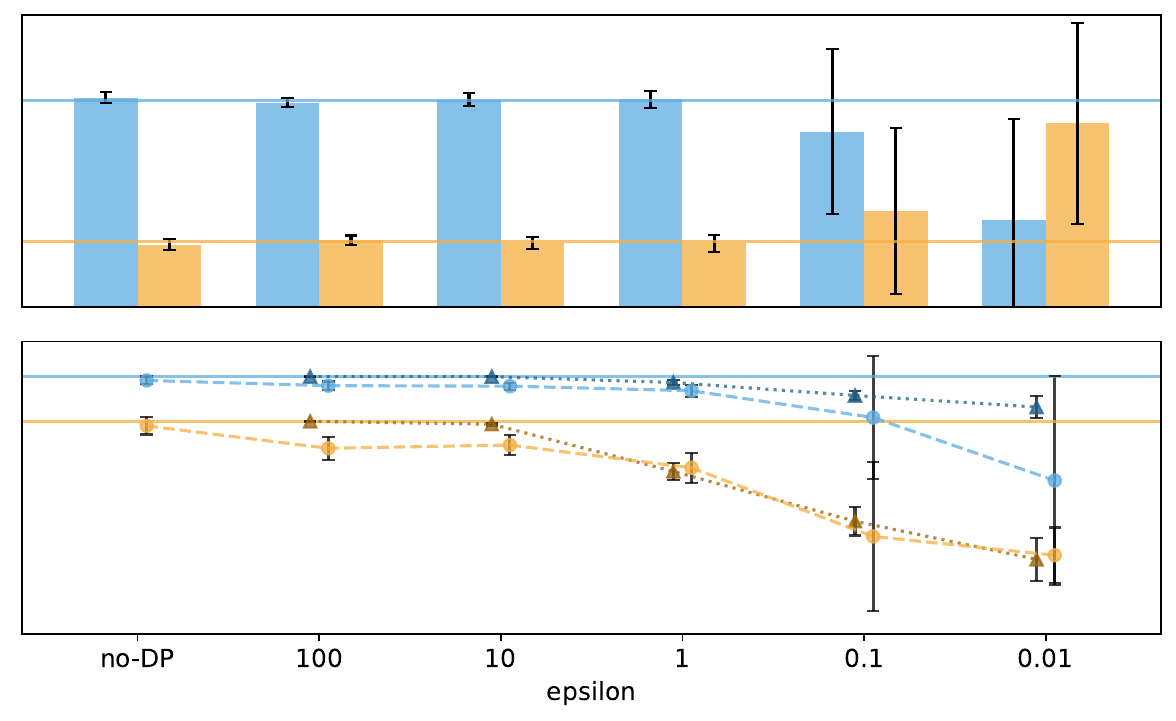}\label{fig:CSPR_A_DPWGAN}}\hspace*{-0.175cm}
	\subfigure[\scriptsize PATE-GAN]{\includegraphics[width=0.34\textwidth]{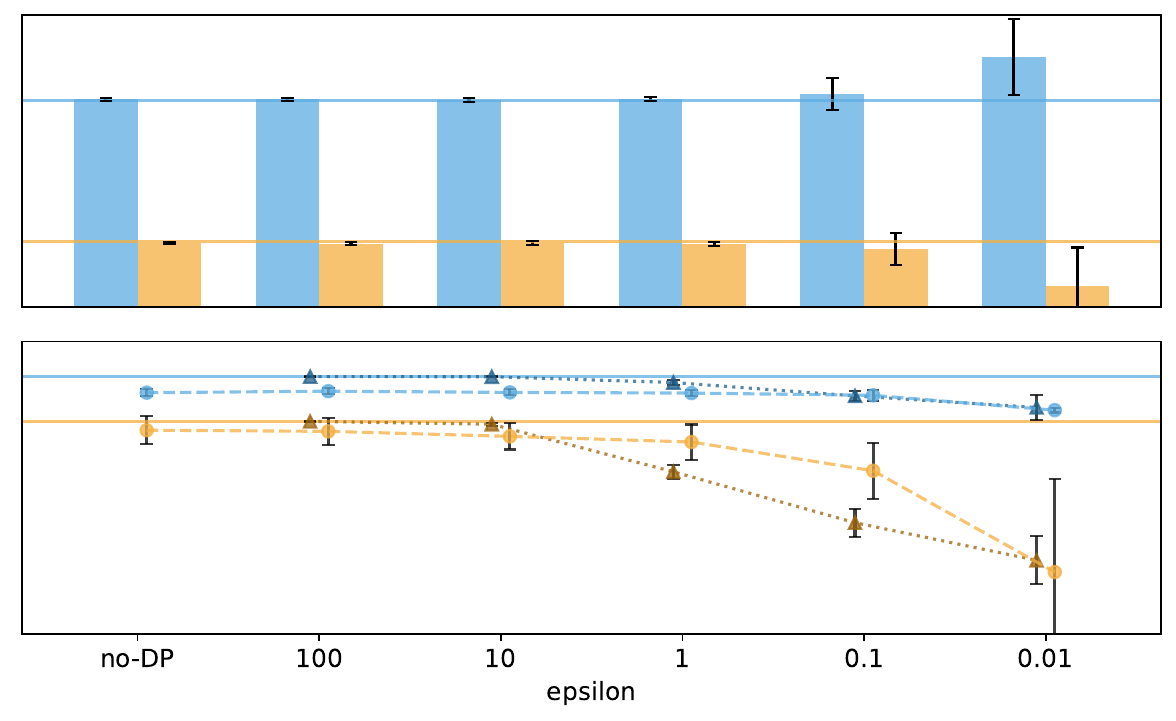}\label{fig:CSPR_A_PATEGAN}}
	\caption{Synthetic data class size (top) and real, DP, and synthetic classifiers precision (bottom) for different levels of $\epsilon$, {\bf\em Adult}, ({\bf\em S1}).}
	\label{fig:CSPR_A}
\end{figure*}

\subsection{S1: Binary Class Size and Precision}
\label{ssec:CSPR}

We consider privacy budgets ($\epsilon$) of 0.01, 0.1, 1, 10, 100, and infinity (``no-DP'') for the binary classification datasets (Adult and Texas).
We do not imbalance the data because all datasets already have imbalanced classes -- specifically, the proportion of the minority class to the total number of records is 0.24 in Adult and 0.195 in Texas.

\descr{Size.}
In the first row of Fig.~\ref{fig:CSPR_A} (and~\ref{fig:CSPR_T} in Appendix~\ref{app:s1}), we plot the class size distribution in the synthetic data for Adult and Texas, respectively.
For both datasets, for PrivBayes, we observe that decreasing $\epsilon$ results in synthetic data with reduced class imbalance; for PATE-GAN, the opposite is true -- decreasing $\epsilon$ leads to increased class imbalance (except for $\epsilon=0.01$ for Texas).
These results are consistent with the disparate effects from applying Laplace to DP statistics~\cite{kuppam2019fair} as well as PATE to DP neural networks classifiers~\cite{uniyal2021dp}.
Interestingly, DP-WGAN preserves the imbalance for $\epsilon>0.1$.
As expected, there is an increased standard deviation for smaller values of $\epsilon$ for all synthetic datasets.

\descr{Precision.}
In the bottom rows of Fig.~\ref{fig:CSPR_A} (and~\ref{fig:CSPR_T} in Appendix~\ref{app:s1}) we plot the precision of the real, DP, and synth classifiers on the two datasets (recall plots are also in Appendix~\ref{app:s1}).
For the DP classifiers, we find that precision drops disproportionately more for the underrepresented class when decreasing $\epsilon$ for all datasets.

Synth classifiers follow very similar behavior even with small privacy budgets ($\epsilon<1$), regardless of the direction of class distortion in the synthetic datasets.

\begin{figure*}[ht!]
	\centering
	\subfigure[\scriptsize PrivBayes]{\includegraphics[width=0.34\textwidth]{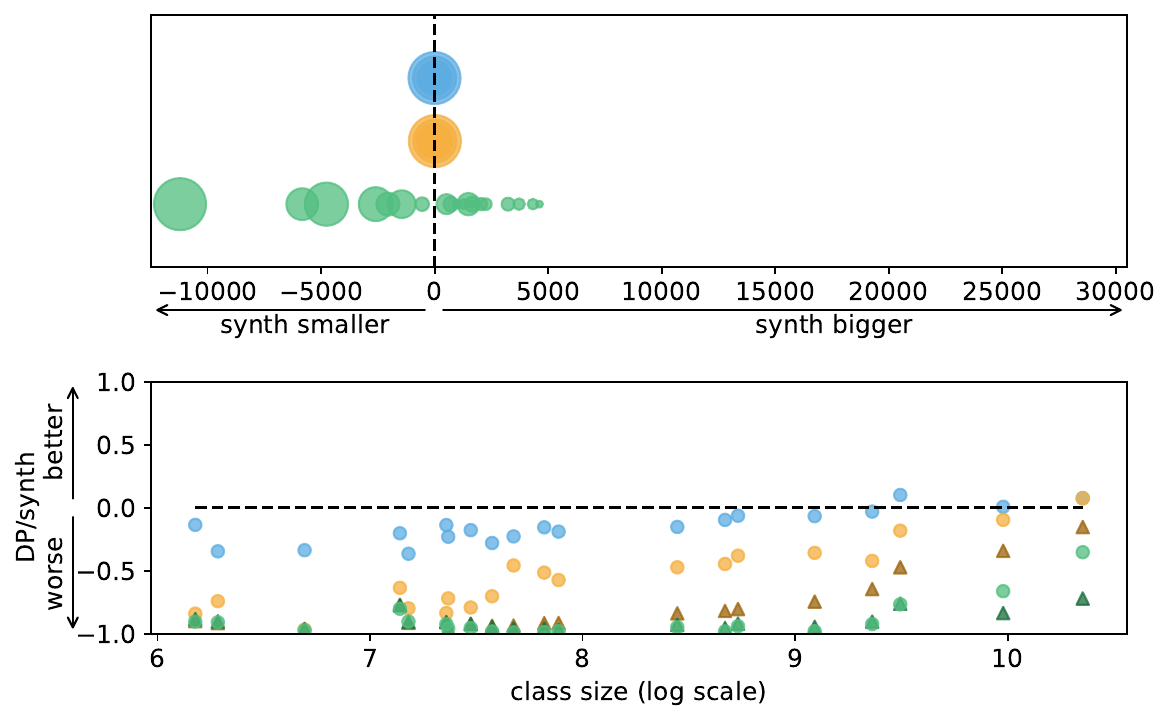}\label{fig:CSPR_P_PrivBayes}}
\hspace*{-0.35cm}
	\subfigure[\scriptsize DP-WGAN]{\includegraphics[width=0.34\textwidth]{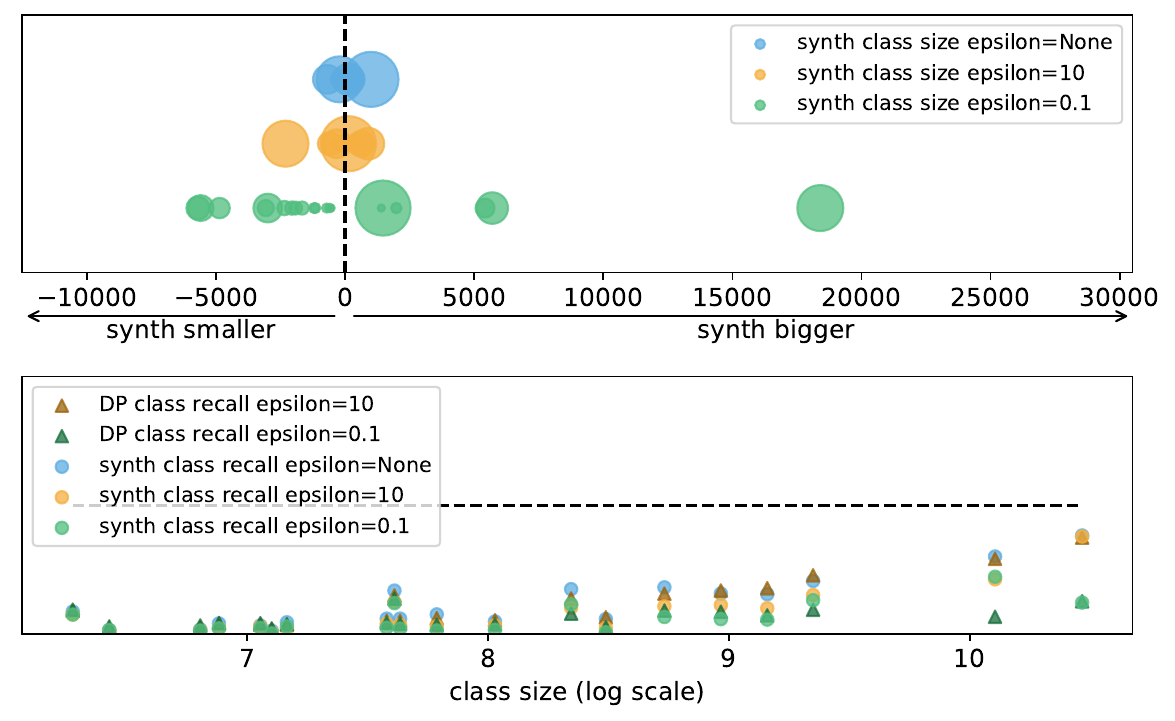}\label{fig:CSPR_P_DPWGAN}}
\hspace*{-0.35cm}
 \subfigure[\scriptsize PATE-GAN]{\includegraphics[width=0.34\textwidth]{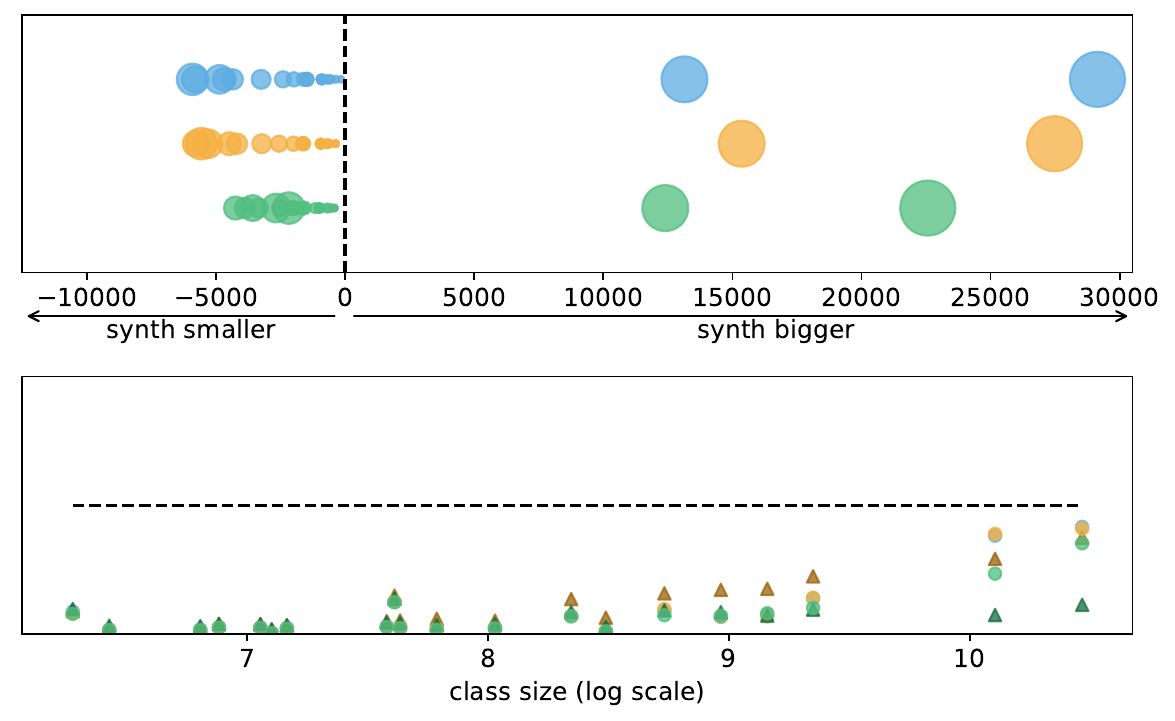}\label{fig:CSPR_P_PATEGAN}}%
	\caption{Synthetic data class (multi-class) size relative to real (top) (each bubble denotes a distinct class while the size its relative count in the real data) and DP and synthetic classifiers recall relative to real (bottom) for different levels of $\epsilon$, {\bf\em Purchases}, ({\bf\em S2}).}
	\label{fig:CSPR_P}
\end{figure*}

\begin{figure*}[t!]
	\centering
	\subfigure{\includegraphics[width=0.99\textwidth]{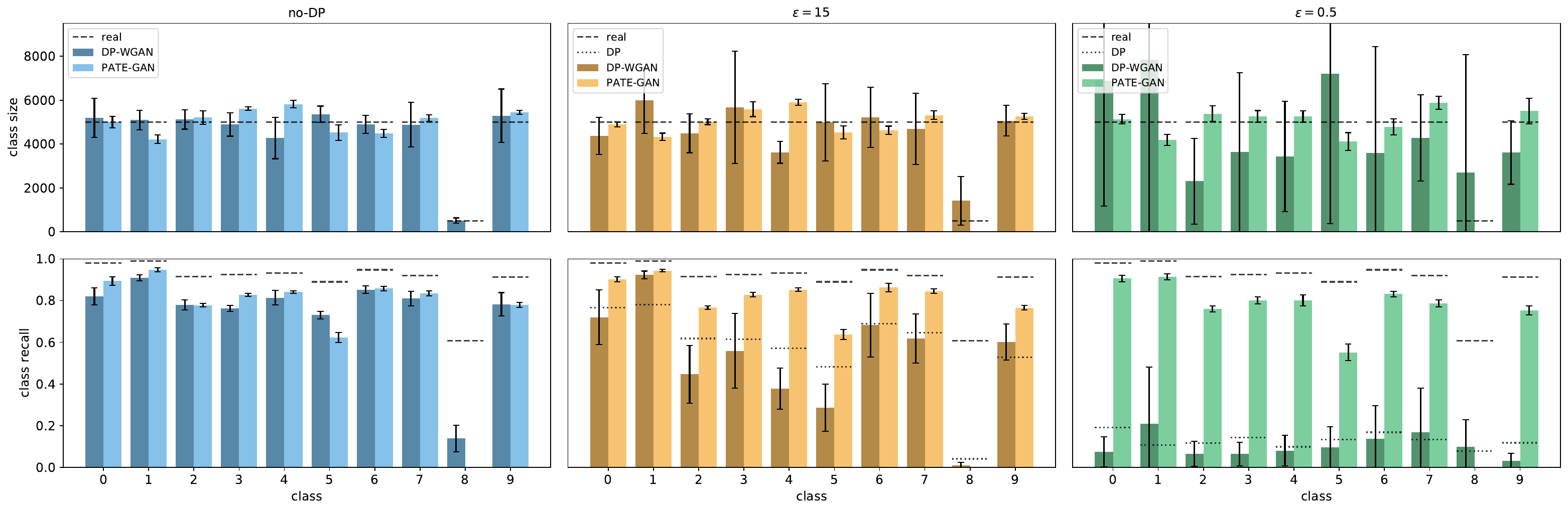}}
	\caption{(S2) Synthetic data class (multi-class) size (top) and real, DP, and synthetic classifiers recall (bottom) for different digits and levels of $\epsilon$, {\bf\em MNIST} with class ``8'' downsampled to 0.1 its count, ({\bf\em S2}).}
	\label{fig:CSPR_M_10}
\end{figure*}

\subsection{S2: Multi-Class Size and Recall}
\label{ssec:mCSPR}
We experiment with privacy budgets ($\epsilon$) of 0.1, 10, and infinity (``no-DP'') for Purchases and 0.5, 5, 15, and infinity (``no-DP'') for MNIST (to maintain consistency with previous work~\cite{uniyal2021dp}).
While for Purchases we do not imbalance the dataset, since it is already imbalanced (the proportion of the smallest to the largest class is 0.015), we imbalance the class ``8'' in MNIST to be 0.25 and 0.1 of the largest class.
Furthermore, for MNIST, we only compare DP-SGD and PATE-GAN because the data has too many dimensions for PrivBayes.

\descr{Size.}
The size of the synthetic data is shown in the top rows of Fig.~\ref{fig:CSPR_P},~\ref{fig:CSPR_M_10}, (and~\ref{fig:CSPR_M_25} in Appendix~\ref{app:mnist}).
For the Purchases dataset, we see similar trends with PrivBayes and PATE-GAN as in S1; the former evens the classes, while the latter increases the gap by ``transferring'' counts from the minority to the majority classes.
PATE-GAN exhibits the strongest disparity, even with ``no-DP,'' which contradicts findings from~\cite{uniyal2021dp}.%
However, the data imbalance here is of different nature as we have two classes with much higher counts than the others rather than a single underrepresented class.
In turn, this could potentially bias the generator towards these classes as the teacher-discriminators are exposed predominantly to them and thus, learn to distinguish them from fake examples better.
DP-WGAN does not preserve the class sizes as successfully as in S1.

For MNIST, PATE-GAN exhibits far better performance for both imbalances and preserves the counts even for lower $\epsilon$ budgets.
Looking at the minority class ``8,'' however, PATE-GAN fails to generate any digits for imbalance 0.1 (even for ``no-DP'' as well).
This could be because the teachers fail to pass samples ``8'' labeled as real to the student even though when applied to classification, PATE is more robust under similar imbalance levels~\cite{uniyal2021dp}.

\descr{Recall.}
In the bottom two rows of Fig.~\ref{fig:CSPR_P},~\ref{fig:CSPR_M_10}, (and~\ref{fig:CSPR_M_25} in Appendix~\ref{app:mnist}), we report the recall of the real, DP, and synth classifiers on the two datasets.
For Purchases, the synth classifiers trained on data from PrivBayes far outperform both the DP classifiers and the other synth classifiers.
Even with ``no-DP,'' the synth classifiers trained on DP-WGAN and PATE-GAN incur a severe recall drop on smaller subgroups.

\begin{figure}[t!]
	\centering
	\hspace*{-0.2cm}
	\subfigure{\includegraphics[width=0.5\textwidth]{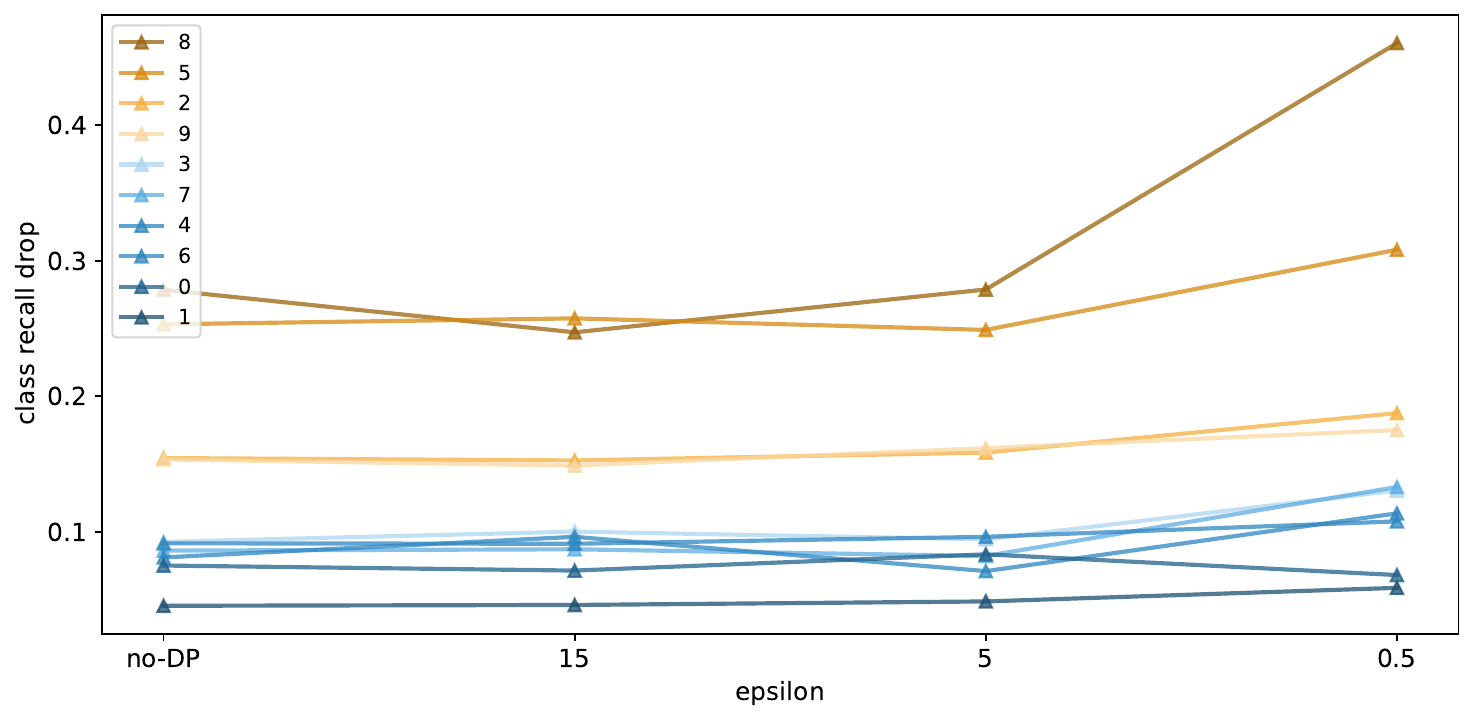}}
	\caption{(S2) PATE-GAN synthetic classifier class recall drop relative to real for different digits and levels of $\epsilon$, {\bf\em MNIST} with class ``8'' downsampled to 0.25 its count, ({\bf\em S2}).}
	\label{fig:mnist_cre}
\end{figure}

For MNIST, again PATE-GAN performs much better than DP-WGAN and DP classifiers -- the recall drops are not so acute, and their standard deviations are much lower.
DP-WGAN follows closely DP classifiers but is slightly worse for all levels of $\epsilon$ and imbalances.
DP-WGAN's performance looks random for $\epsilon=0.5$, which means that the classifiers failed to learn anything, most likely due to bad quality of the synthetic data (Fig.~\ref{fig:mnist_digits} in Appendix~\ref{app:mnist}).
While expectedly DP-WGAN monotonically drops in terms of recall with decreasing $\epsilon$, PATE-GAN's performance actually increases when DP is applied, e.g., $\epsilon=15$ and 5 yield marginally better results than ``no-DP'' for both imbalances.
This is most likely due to the fact that the teacher-discriminators are exposed to different subsets of the real data, and as result, do not learn exactly the same distributions as well as the noise added to their votes, which further enables generalization.
Interestingly, the performance on some digits suffers a lot more than others (e.g., ``2,'' ``5,'' ``9''), which could be explained because they are visually close to ``8.''
This phenomenon is displayed in Fig.~\ref{fig:mnist_cre}.
The observation allows us to speculate that applying DP could not only lead to worse performance for the underrepresented subparts of the data but also for those with similar characteristics as well.

\subsection{S3: Single-Attribute Subgroup Size, Accuracy, and Correlation}
\label{ssec:SSA}
We treat a single feature -- namely, sex -- as a subgroup in the Adult and Texas datasets.
We consider privacy budgets ($\epsilon$) of 0.01, 0.1, 1, 10, 100, and infinity (``no-DP'') as well as imbalance ratios of 0.01, 0.05, 0.1, 0.25, and 0.5.

\begin{figure}[t!]
	\centering
	\hspace*{-0.2cm}
	\subfigure{\includegraphics[width=0.5\textwidth]{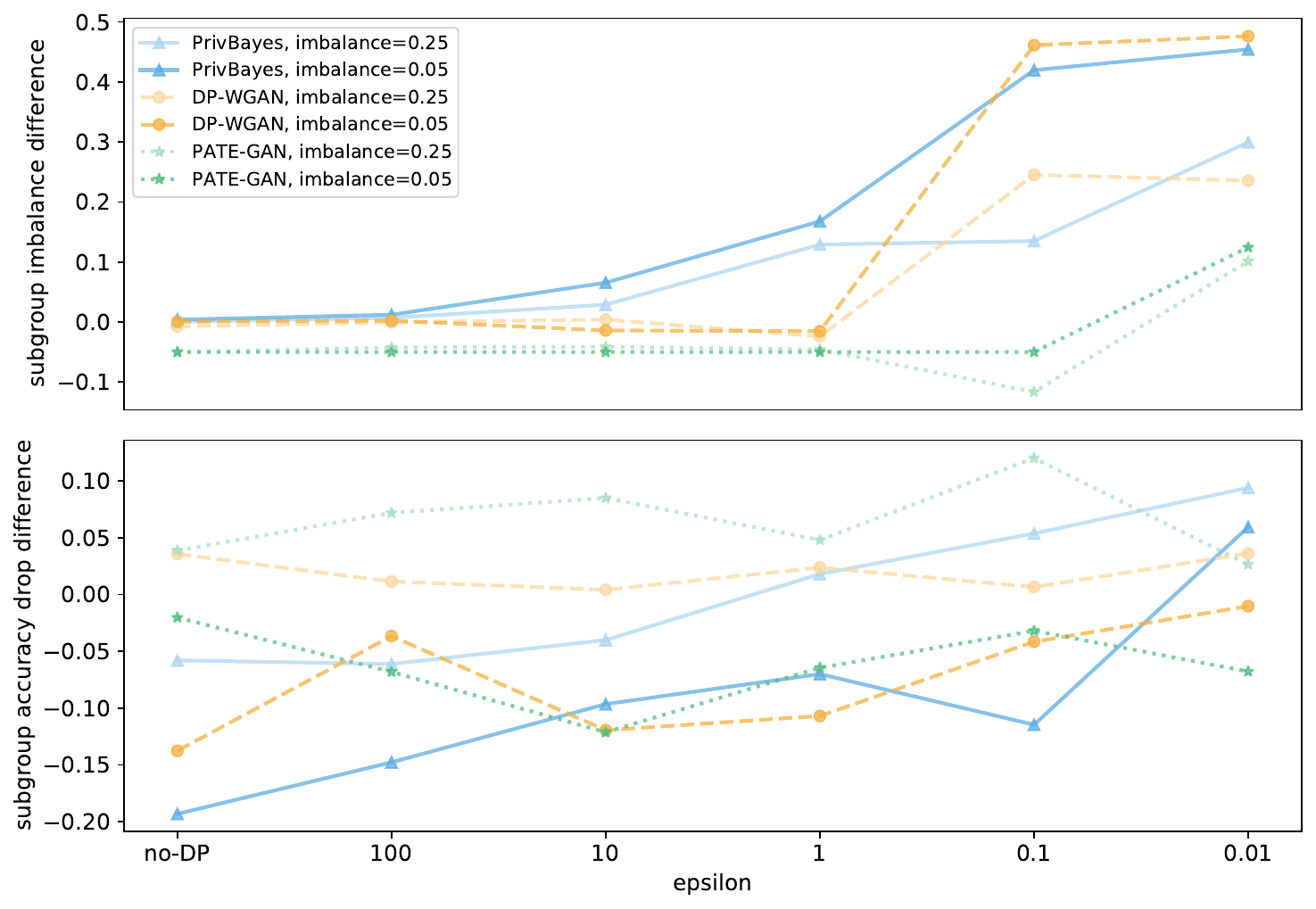}}
	\caption{Minority single-attribute (sex) subgroup imbalance level difference (top) and minority subgroup accuracy drop difference (bottom) relative to majority for different subgroup imbalance and $\epsilon$ levels, {\bf\em Texas}, ({\bf\em S3}).}
	\label{fig:s3_T}
\end{figure}

\descr{Size.}
In the top rows of the plots in Fig.~\ref{fig:SSA} in Appendix~\ref{app:s3}, we report the full experiments for $\epsilon$ and imbalance effects on subgroup size on the two datasets while in Fig.~\ref{fig:s3_T} we summarize the trends for Texas.
Once again, we find that, with PrivBayes, decreasing $\epsilon$ results in synthetic data with reduced subgroup imbalance for all datasets -- the higher the initial imbalance, the more PrivBayes balances the subgroups (could be seen in the slope of the blue lines).
As before, this effect could be attributed to the truncation of negative noisy counts after the Laplace mechanism is applied.
PATE-GAN synthetic datasets follow the opposite trend.
The gap becomes so large that, for imbalances lower than 0.1 for Adult and 0.25 for Texas, PATE-GAN barely generates the underrepresented subgroup for all $\epsilon$ values except 0.01.
DP-WGAN is again the most successful at preserving the imbalance for $\epsilon>0.1$.
For $\epsilon=0.1$ and 0.01, the subgroup size appears random, as the DP-WGAN models are trained only for a few iterations before the full privacy budget is spent.

\descr{Accuracy.}
The bottom rows of %
Fig.~\ref{fig:SSA} in Appendix~\ref{app:s3} %
report the accuracy of the various classifiers, while Fig.~\ref{fig:s3_T} displays a summary for Texas.
Interestingly, we find that the real classifier, with Adult, achieves higher accuracy on the underrepresented subgroup ``Female'' than the overrepresented ``Male'' for all imbalances.

As for DP classifiers, over a certain $\epsilon$, decreasing it further reduces the accuracy of the minority subgroup more than that of the majority.
Additionally, this reduction in accuracy is more accentuated with increasing subgroup imbalance.
For example, looking at the Adult plots, the accuracy on ``Female'' drops more than ``Male'' for $\epsilon\leq0.1$ and imbalance 0.5.
For increasing imbalances, the drop on the minority subgroup overtakes the majority for larger privacy budgets, i.e., $\epsilon\leq1$ for imbalances 0.25 and 0.1, $\epsilon\leq10$ for imbalance 0.05, and finally $\epsilon\leq100$ for imbalance 0.01.

The synth classifiers incur a bigger accuracy drop in the underrepresented subgroup---regardless of the subgroup sizes in the synthetic data (as observed in the overall positive slopes of all lines in the bottom row of Fig.~\ref{fig:s3_T}).
Classifiers trained on PrivBayes synthetic data follow the behavior of DP classifiers the closest.
Overall, DP-WGAN synth classifiers perform worse than the others as they are more unstable, with some noticeable accuracy drops (e.g., $\epsilon=100$ for imbalance 0.5 in Adult, $\epsilon=1$ for imbalances 0.1 and 0.05 in Texas).
PATE-GAN synth classifiers have better accuracy than DP classifiers for both subgroups for $\epsilon<10$; this is perhaps surprising, especially for the underrepresented subgroup, as the synth classifiers are trained on synthetic data containing only a small number of the minority subgroup.

\descr{Correlation.}
The mutual information between the subgroup and the target columns are displayed in Fig.~\ref{fig:sS_MI} in Appendix~\ref{app:s3}.
Since in the preparation step the subgroups were balanced by class, the expected (and the real) mutual information is 0.
We observe that, DP-WGAN manages best to maintain this relationship, closely followed by PrivBayes which, however, has a few noisy exceptions for small privacy budgets $\epsilon<=0.1$.
In contrast, PATE-GAN introduces undesirable artifact in the synthetic data for $\epsilon<=0.1$ for both datasets.
In other words, the model creates data with stronger relationship between the subgroup column and the target column.

\begin{figure}[t!]
	\centering
	\subfigure[\scriptsize PrivBayes]{\includegraphics[width=0.495\textwidth]{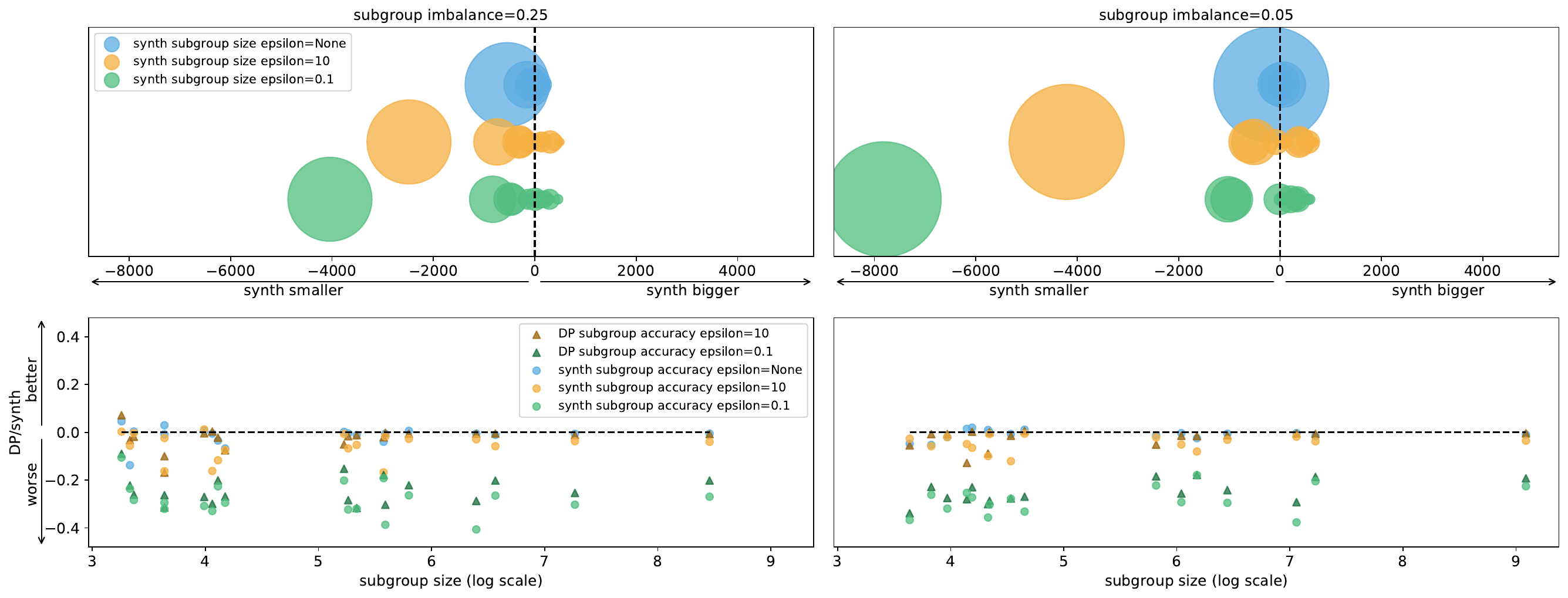}\label{fig:SsSA_A_PrivBayes}}\\[-1ex]
	\subfigure[\scriptsize DP-WGAN]{\includegraphics[width=0.495\textwidth]{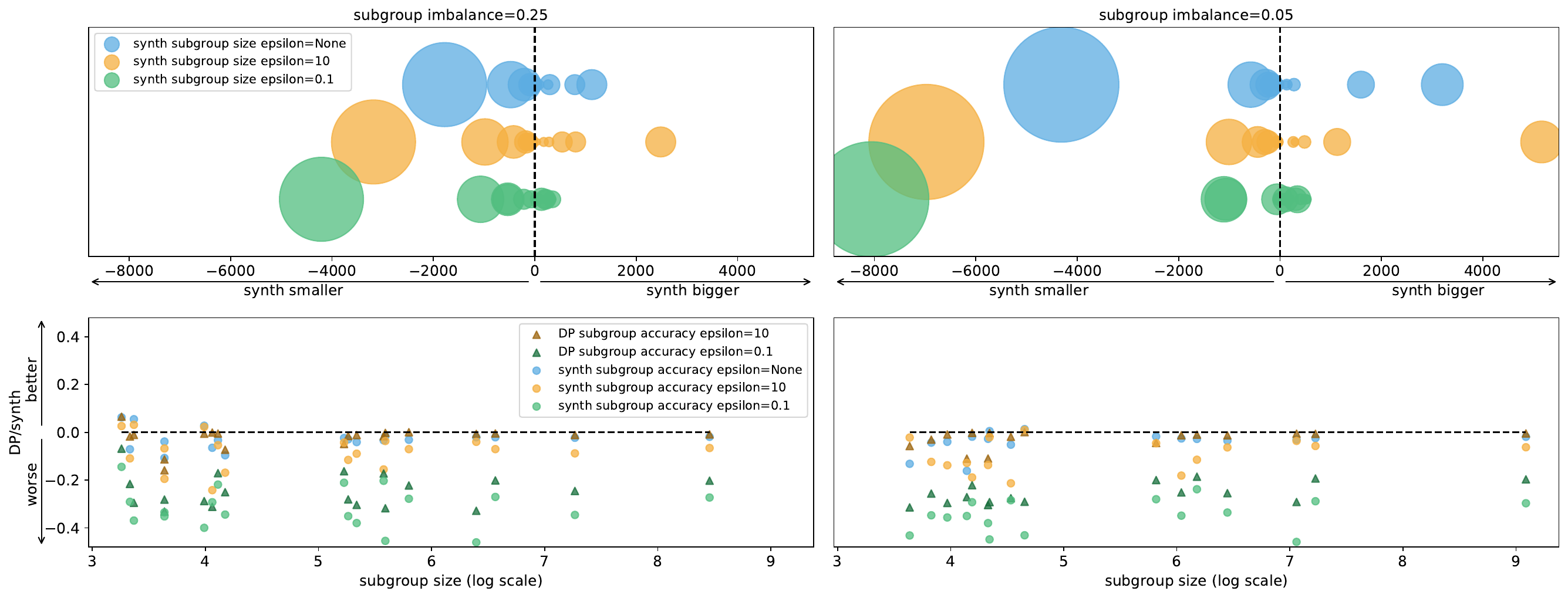}\label{fig:SsSA_A_DPWGAN}}\\[-1ex]
	\subfigure[\scriptsize PATE-GAN]{\includegraphics[width=0.495\textwidth]{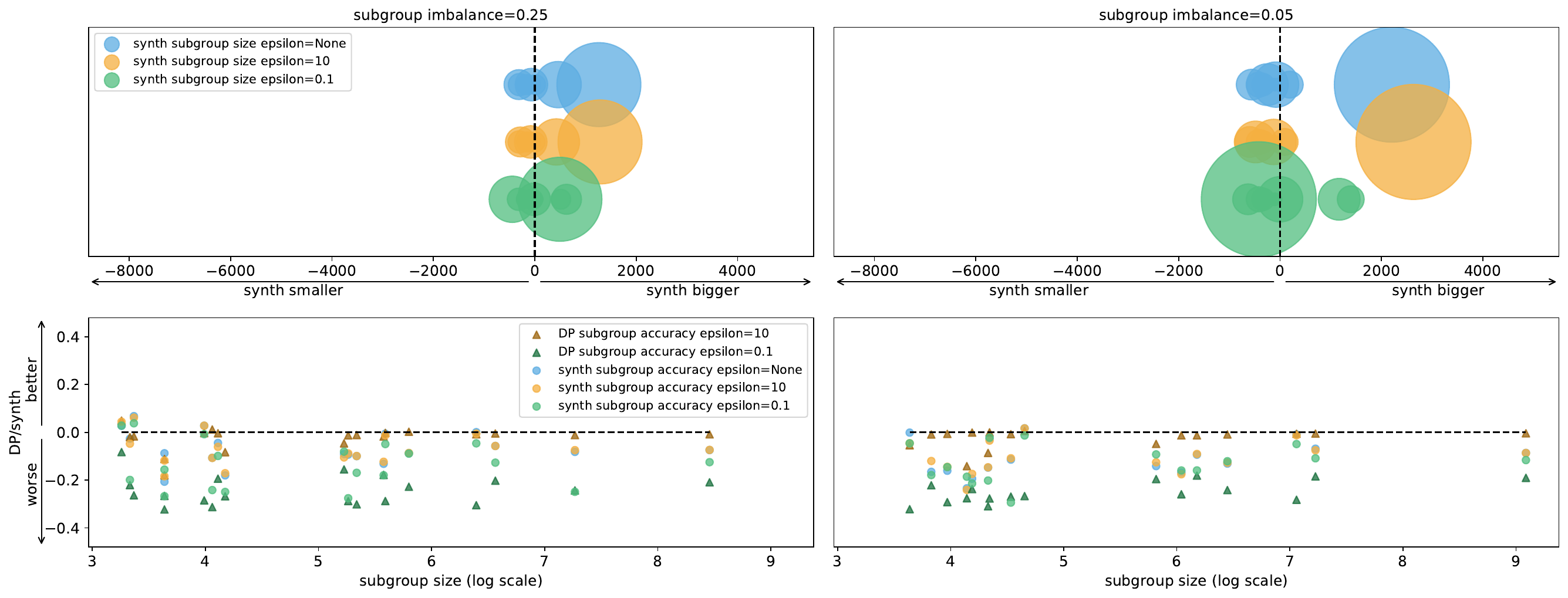}\label{fig:SsSA_A_PATEGAN}}
	\caption{Synthetic data multi-attribute (intersection of age, sex, and race) subgroup size relative to real (top) (each bubble denotes a distinct subgroup while the size its relative count in the real data) and DP and synthetic classifiers accuracy relative to real (bottom) for different single-attribute (sex) subgroup imbalance and $\epsilon$ levels, {\bf\em Adult}, ({\bf\em S4}).}
	\label{fig:SsSA_A}
\end{figure}

\begin{figure}[t!]
	\centering
	\subfigure[\scriptsize PrivBayes, Adult]{\includegraphics[width=0.495\textwidth]{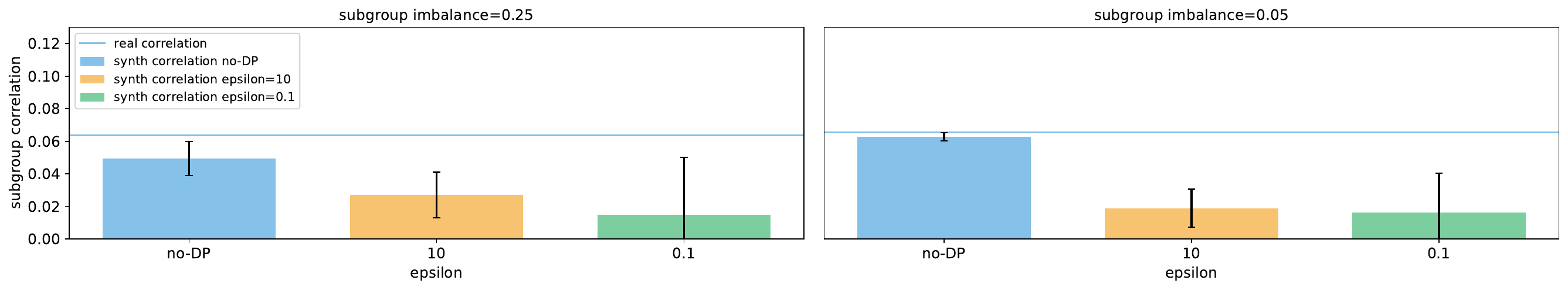}}\\[-1ex]
	\subfigure[\scriptsize DP-WGAN, Adult]{\includegraphics[width=0.495\textwidth]{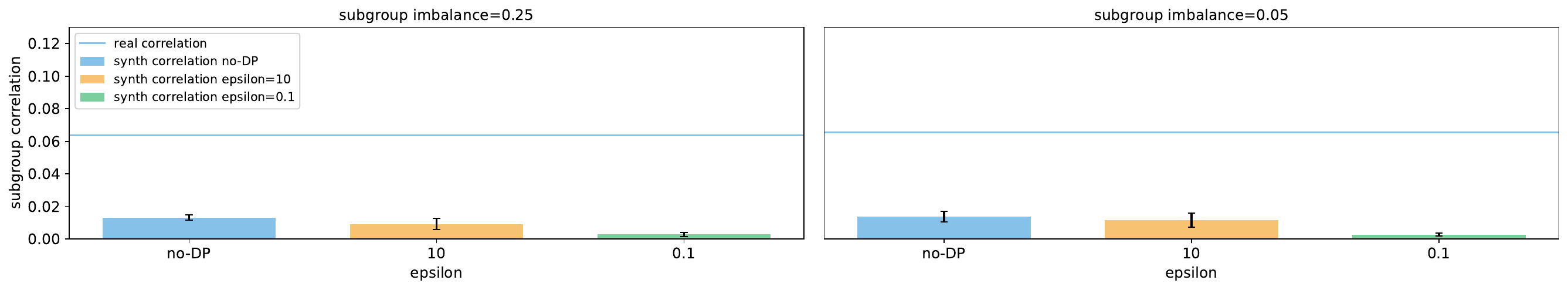}}\\[-1ex]
	\subfigure[\scriptsize PATE-GAN, Adult]{\includegraphics[width=0.495\textwidth]{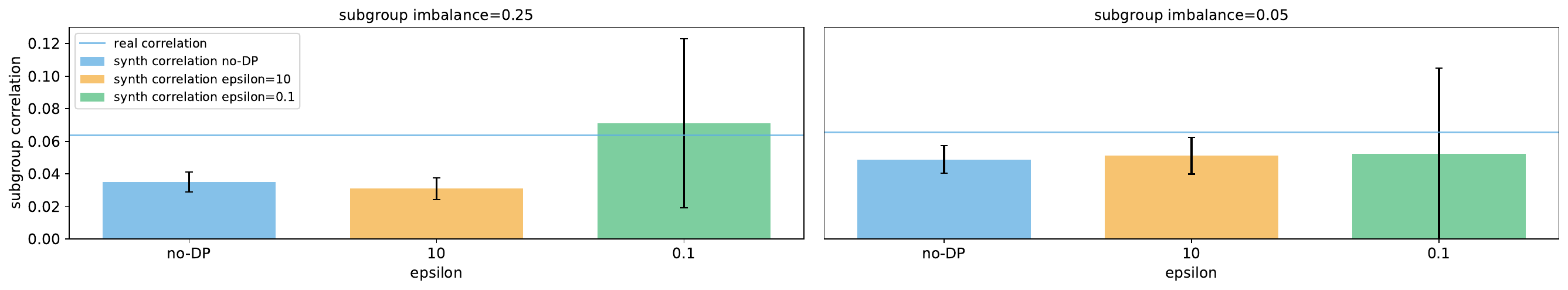}}\\[-1ex]
	\subfigure[\scriptsize PrivBayes, Texas]{\includegraphics[width=0.495\textwidth]{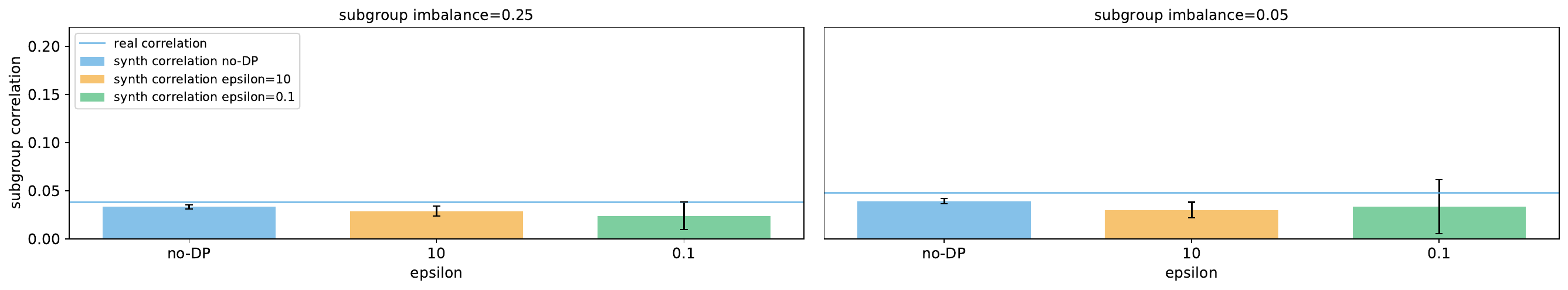}}\\[-1ex]
	\subfigure[\scriptsize DP-WGAN, Texas]{\includegraphics[width=0.495\textwidth]{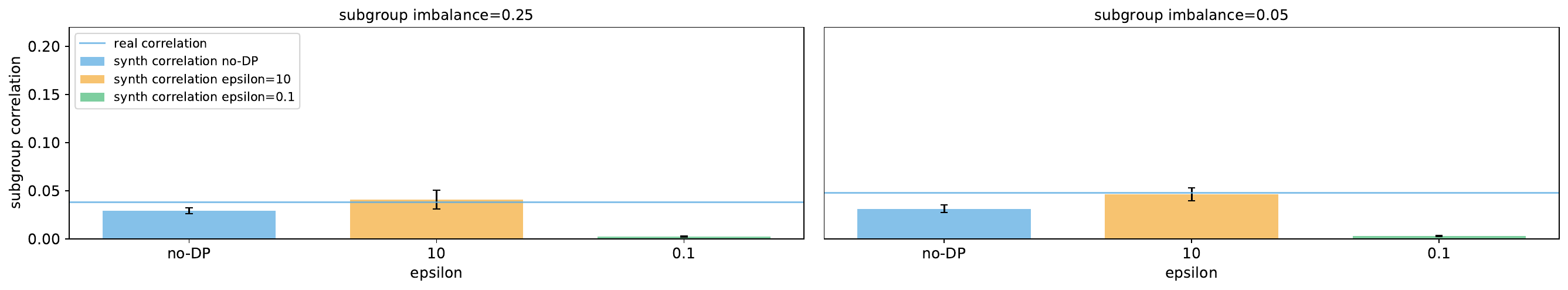}}\\[-1ex]
	\subfigure[\scriptsize PATE-GAN, Texas]{\includegraphics[width=0.495\textwidth]{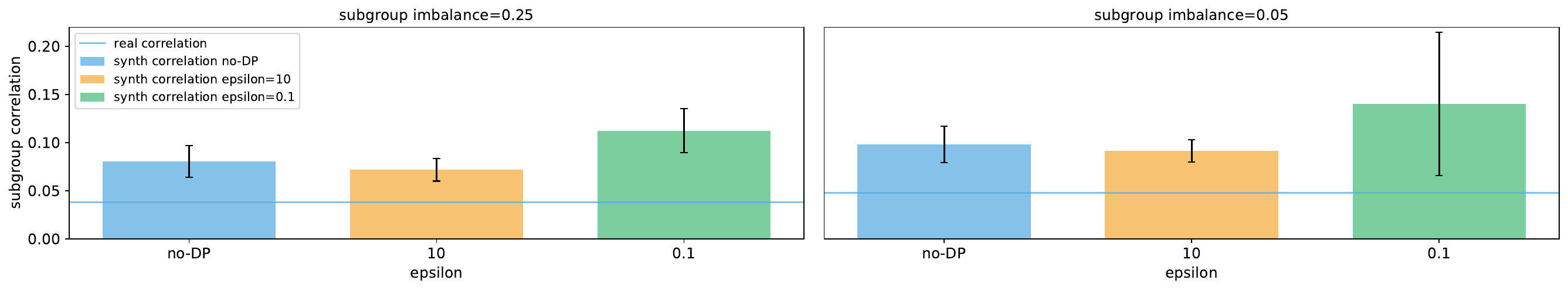}}
	\caption{Mutual information between the multi-attribute (intersection of age, sex, and race) subgroup and the target (income/length of stay) columns for different single-attribute subgroup imbalance (sex) and $\epsilon$ levels, {\bf\em Adult} (top 3) and {\bf\em Texas} (bottom 3), ({\bf\em S4}).}
	\label{fig:mS_MI}
\end{figure}

\subsection{S4: Multi-Attribute Subgroup Size, Accuracy, and Correlation}
\label{ssec:SsSA}
In our last set of experiments, we treat the intersection between three features -- age, sex, and race -- as complex subgroups in the Adult and Texas datasets.
We consider privacy budgets ($\epsilon$) of 0.1, 10, and infinity (``no-DP'') as well as imbalance ratios of 0.05 and 0.25.
This results in 16 subgroups for Adult, 21 for Texas for imbalance 0.05 and 19, and 27 for 0.25, respectively.

\descr{Size.}
In the top rows of the plots in Fig.~\ref{fig:SsSA_A} (also see Fig.~\ref{fig:SsSA_T} in Appendix~\ref{app:s4texas}), we report the sizes of the subgroups, with the three different models and the two datasets.
Once again, PrivBayes reduces the gap between majority and minority subgroups in the synthetic data, whereas PATE-GAN increases it.
DP-WGAN behaves similarly to PrivBayes, which is inconsistent with the multi-class case discussed in Sec.~\ref{ssec:SSA}.
For Adult, DP-WGAN does not manage to keep the subgroups distribution, even for ``no-DP.''
Finally, the effect of increased subgroup imbalance is evident in the higher disparity for all models and datasets.

\descr{Accuracy.}
The bottom rows of the plots in Fig.~\ref{fig:SsSA_A} (and~\ref{fig:SsSA_T}) report the performance of the classifiers.
In contrast to previous work~\cite{bagdasaryan2019differential}, we do not observe ``the rich get richer, the poor get poorer'' effect for either DP or synth classifiers; instead, ``everybody gets poorer,'' with very few exceptions.
This should not come as a surprise, as it could not be expected from a generative model to synthesize data with better utility than the real data.
The mentioned exceptions are only for subgroups with small sizes; however, the accuracy on these subgroups has a much higher standard deviation compared to larger subgroups.

There is also a clear distinction between classifiers with $\epsilon=10$ and 0.1, both DP and synth.
Synth classifiers trained on PATE-GAN and PrivBayes synthetic data incur a smaller drop than DP-WGAN synthetic data.
Similar to S3, PATE-GAN trained synth classifiers have better accuracy than DP classifiers for $\epsilon=0.1$.

\descr{Correlation.}
In Fig.~\ref{fig:mS_MI}, we display the mutual information between the multi-attribute subgroup and the target.
Unlike the single-attribute scenario, the baseline mutual information here is not 0 because only one of the attributes (sex) was balanced by class.
For both datasets, PrivBayes exhibits the most expected behavior -- increasing the privacy budget $\epsilon$ results in more distorted synthetic data, thus reducing the mutual information between the subgroup and target columns.
On the other hand, PATE-GAN displays similar to u-shaped behavior: incorporating some privacy ($\epsilon=10$) initially reduces the mutual information but adding more privacy ($\epsilon=0.1$) increases it to level even higher than when ``no-DP'' is applied.
In particular, for the Texas dataset, PATE-GAN enforces the dependency between the subgroups and target columns for all privacy budgets.
Finally, DP-WGAN performs the worst in the Adult dataset.

\subsection{Main Take-Aways}

\descr{Disparate Effects.} Overall, our experiments provide an empirical demonstration that DP generative models {\em do} have different disparate effects on synthetic data.
Analyzing the size of the classes and subgroups in the generated data, we consistently observe that PrivBayes reduces the gap between the majority and minority classes/subgroups (thus exhibiting a ``Robin Hood'' effect), PATE-GAN increases it (exhibiting a ``Matthew'' effect), and DP-WGAN has a mixed one.

\descr{Downstream Classification.} When performing classification tasks on data produced by generative models, one faces an even bigger (or more variable) accuracy drop on minority classes and subgroups.
We also see that higher privacy guarantees and more imbalanced datasets result in more substantial disparate effects.
For example, even though PATE-GAN displays better overall behavior than DP-WGAN, an imbalance of 0.1 prevents it from learning an entire class of the data even for low privacy settings.
High privacy guarantees also result in PATE-GAN generating undesirable artifacts in the synthetic data in the form of a much stronger correlation between low correlated columns.

\descr{Revisiting Our Research Questions.} Recall from Section~\ref{sec:intro} that our work aimed to answer three main research questions; we now summarize some concise answers.

\descrit{RQ1: Do DP generative models generate data in similar classes and subgroups proportions to the real data?}\\
Not really.
DP distorts the proportions, yielding Robin Hood vs Matthew effects depending on the DP generative model.

\descrit{RQ2: Does training a classifier on DP synthetic data lead to the same disparate impact on accuracy as training a DP classifier on the real data?}\\
Overall, yes.
Smaller classes/subgroups suffer more similarly to DP classifiers.
However, we do not see the rich get richer, the poor get poorer; everybody gets poorer.
Incidentally, sometimes synthetic classifiers are better than DP classifiers; studying this in detail is left to future work.

\descrit{RQ3: Do different DP mechanisms for DP synthetic data behave similarly under different privacy and data imbalance levels?}\\
No, different DP generative models behave differently.
For example, PATE-GAN performs better than DP-WGAN, with some very specific exceptions, while PrivBayes is the only one that manages to maintain the data utility for the multi-class tabular data Purchases.

\section{Related Work}
\citet{kuppam2019fair} show that, if resource allocation is decided based on DP statistics, smaller districts could get more funding and larger ones less. %
Prior work has also studied deep neural network classifiers trained using DP-SGD~\cite{bagdasaryan2019differential, farrand2020neither, suriyakumar2021chasing} on imbalanced datasets (mainly images).
Essentially, they show that underrepresented groups in a dataset %
that already incur lower accuracy end up losing even more accuracy when DP is applied.
In particular, %
\citet{farrand2020neither} show that even small imbalances and loose privacy guarantees can cause disparate impacts.

\citet{uniyal2021dp} find that classifiers trained with PATE exhibit disparate drops in performance but less severely than with DP-SGD.
Furthermore, %
\citet{feldman2020does} formalizes the need for accurate discriminative models to memorize training data and studies the disparate effects of privacy and model compression on subgroups.
\citet{chen2020understanding} provide a theoretical analysis that quantifies the clipping bias on convergence with a disparity measure between the gradient and a geometrically symmetric distribution.
There are also papers focusing on learning DP classifiers with fairness constraints~\cite{jagielski2019differentially, tran2021differentially} and on analyzing the PATE framework from a fairness point of view~\cite{tran2021fairness}.
Unlike our work, these efforts focus on discriminative (rather than generative) models.

\citet{cheng2021can} show that training classifiers on DP synthetic images can result in significant utility degradation and increased majority subgroup influence, but not worse group unfairness measures.
However, they only use a single generative model and only look at the utility of balanced DP synthetic datasets.

Finally, recent work has focused on tabular DP synthetic data.
\citet{ghalebikesabi2021bias} introduce novel bias mitigation techniques, which, unfortunately, lead to reduced usefulness of the synthetic data.
Perhaps closer to our work is that by~\citet{pereira2021analysis}, who mainly look at single-attribute subgroup fairness and overall classification performance.
Their work, however, does not investigate the utility disparity on different single and multi-attribute subgroups of the data, nor the effect of data imbalance.
Furthermore, we consider the size disparities in the generated synthetic data as we do not use conditional generative models.
We also experiment with a far wider range of epsilon budgets.
Overall, we are the first to highlight and analyze the disparate effects of several factors: generative model type, DP mechanism, privacy budget, class and single/multi-attribute subgroup imbalance on the resulting synthetic data in terms of both statistical analysis and downstream classification.
We do so through experiments geared to isolate the effect of these factors.

There is a rich literature with DP generative models for tabular data~\cite{acs2018differentially, xie2018differentially, tantipongpipat2019differentially, frigerio2019differentially, zhang2021privsyn, mckenna2021winning}.
Our goal is, however, not to benchmark all possible models but to focus on the best known and accessible state-of-the-art models relying on different/well-studied DP mechanisms (Laplace, DP-SGD, PATE).

\section{Conclusion}
This work analyzed the effects of privacy-preserving generative models, using different DP methods,
on 1) class/subgroups distributions in the generated synthetic data and 2) the performance of downstream tasks.
We found that applying DP to synthetic data generation disparately affects the minority subpopulations.
As for the class/subgroup distribution in the synthetic data, DP can have opposing effects depending on the underlying DP method; e.g., PrivBayes reduces the imbalance, PATE-GAN increases it.
However, when training a classifier on the synthetic data, minority subpopulations suffer stronger and/or more varying decreases in accuracy.
We also showed that the privacy budget and data imbalance are important factors and further intensify these effects.

Overall, our work %
motivates the need for practitioners and companies to take the disparate effects into consideration and adopt more extensive testing before deploying synthetic data, given that the studied technologies are already in production in the real world~\cite{synth2020, dpcensus2021} and there are concerns and scepticism from the public~\cite{changes2020, census2020}.

We are confident that our results will motivate further research (including theoretical contributions) at the intersection of generative models, DP, and fairness.
Hopefully, this will include novel generative models with modified/new DP learning algorithms that could reproduce the original data in a privacy-preserving manner and without disparate loss in utility.
Another interesting direction would be to examine the conditions under which classifiers trained on DP synthetic data achieve better utility than DP classifiers trained on real data as observed in this paper.

To facilitate further research in this space, including reproducibility of our results and analysis a wider set of hyperparameters, additional generative models, datasets, and/or tasks beyond classification, we are also open-sourcing our code.\footnote{\url{https://github.com/ganevgv/dp-gen-disparate}}
As part of future work, we plan to explore the relationship between disparate effects and fairness, support additional experiments, as well as release a re-usable, modular framework that integrates with other security and privacy evaluations of both discriminative and generative models~\cite{liu2021ml, stadler2022synthetic}.

\let\OLDthebibliography\thebibliography
\renewcommand\thebibliography[1]{
  \OLDthebibliography{#1}
  \setlength{\parskip}{2pt}
  \setlength{\itemsep}{2pt plus 0.3ex}
}

{\small
\bibliography{mybib}

\begin{thebibliography}{59}
\providecommand{\natexlab}[1]{#1}
\providecommand{\url}[1]{\texttt{#1}}
\expandafter\ifx\csname urlstyle\endcsname\relax
  \providecommand{\doi}[1]{doi: #1}\else
  \providecommand{\doi}{doi: \begingroup \urlstyle{rm}\Url}\fi

\bibitem[Abadi et~al.(2016)Abadi, Chu, Goodfellow, McMahan, Mironov, Talwar,
  and Zhang]{abadi2016deep}
Abadi, M., Chu, A., Goodfellow, I., McMahan, H.~B., Mironov, I., Talwar, K.,
  and Zhang, L.
\newblock {Deep learning with differential privacy}.
\newblock In \emph{{ACM CCS}}, 2016.

\bibitem[Acs et~al.(2018)Acs, Melis, Castelluccia, and
  De~Cristofaro]{acs2018differentially}
Acs, G., Melis, L., Castelluccia, C., and De~Cristofaro, E.
\newblock {Differentially private mixture of generative neural networks}.
\newblock \emph{IEEE TKDE}, 2018.

\bibitem[Alzantot \& Srivastava(2019)Alzantot and
  Srivastava]{alzantot2019differential}
Alzantot, M. and Srivastava, M.
\newblock {Differential Privacy Synthetic Data Generation using WGANs}.
\newblock
  \url{https://github.com/nesl/nist_differential_privacy_synthetic_data_challenge/},
  2019.

\bibitem[Arjovsky et~al.(2017)Arjovsky, Chintala, and
  Bottou]{arjovsky2017wasserstein}
Arjovsky, M., Chintala, S., and Bottou, L.
\newblock {Wasserstein generative adversarial networks}.
\newblock In \emph{{ICML}}, 2017.

\bibitem[Bagdasaryan et~al.(2019)Bagdasaryan, Poursaeed, and
  Shmatikov]{bagdasaryan2019differential}
Bagdasaryan, E., Poursaeed, O., and Shmatikov, V.
\newblock {Differential privacy has disparate impact on model accuracy}.
\newblock In \emph{NeurIPS}, 2019.

\bibitem[Barber(2012)]{barber2012bayesian}
Barber, D.
\newblock \emph{{Bayesian reasoning and machine learning}}.
\newblock Cambridge University Press, 2012.

\bibitem[Benedetto et~al.(2018)Benedetto, Stanley, Totty,
  et~al.]{benedetto2018creation}
Benedetto, G., Stanley, J.~C., Totty, E., et~al.
\newblock {The creation and use of the SIPP synthetic Beta v7. 0}.
\newblock \emph{US Census Bureau}, 2018.

\bibitem[Brown(2020)]{synth2020}
Brown, A.
\newblock {Synthetic Data Promises Fair AI And Privacy Compliance, But How
  Exactly Does It Work?}
\newblock \url{https://tinyurl.com/yc5vtrhb}, 2020.

\bibitem[Carlini et~al.(2019)Carlini, Liu, Erlingsson, Kos, and
  Song]{carlini2019secret}
Carlini, N., Liu, C., Erlingsson, {\'U}., Kos, J., and Song, D.
\newblock {The secret sharer: Evaluating and testing unintended memorization in
  neural networks}.
\newblock In \emph{USENIX Security}, 2019.

\bibitem[Chaudhuri et~al.(2011)Chaudhuri, Monteleoni, and
  Sarwate]{chaudhuri2011differentially}
Chaudhuri, K., Monteleoni, C., and Sarwate, A.~D.
\newblock {Differentially private empirical risk minimization}.
\newblock \emph{JMLR}, 2011.

\bibitem[Chen et~al.(2020{\natexlab{a}})Chen, Yu, Zhang, and
  Fritz]{chen2020gan}
Chen, D., Yu, N., Zhang, Y., and Fritz, M.
\newblock {Gan-leaks: A taxonomy of membership inference attacks against
  generative models}.
\newblock In \emph{ACM CCS}, 2020{\natexlab{a}}.

\bibitem[Chen et~al.(2020{\natexlab{b}})Chen, Wu, and
  Hong]{chen2020understanding}
Chen, X., Wu, S.~Z., and Hong, M.
\newblock {Understanding gradient clipping in private SGD: A geometric
  perspective}.
\newblock \emph{NeurIPS}, 2020{\natexlab{b}}.

\bibitem[Cheng et~al.(2021)Cheng, Suriyakumar, Dullerud, Joshi, and
  Ghassemi]{cheng2021can}
Cheng, V., Suriyakumar, V.~M., Dullerud, N., Joshi, S., and Ghassemi, M.
\newblock {Can You Fake It Until You Make It? Impacts of Differentially Private
  Synthetic Data on Downstream Classification Fairness}.
\newblock In \emph{ACM FAccT}, 2021.

\bibitem[DSHS(2013)]{dshs2013texas}
DSHS.
\newblock {Texas Hospital Inpatient Discharge Public Use Data File Q1-Q4,
  2013}.
\newblock \url{https://www.dshs.texas.gov/THCIC/Hospitals/Download.shtm}, 2013.

\bibitem[Dua \& Graff(2017)Dua and Graff]{dua2017adult}
Dua, D. and Graff, C.
\newblock {UCI Machine Learning Repository}.
\newblock \url{https://archive.ics.uci.edu/ml/datasets/adult}, 2017.

\bibitem[Dwork et~al.(2006{\natexlab{a}})Dwork, Kenthapadi, McSherry, Mironov,
  and Naor]{dwork2006our}
Dwork, C., Kenthapadi, K., McSherry, F., Mironov, I., and Naor, M.
\newblock {Our data, ourselves: Privacy via distributed noise generation}.
\newblock In \emph{EuroCrypt}, 2006{\natexlab{a}}.

\bibitem[Dwork et~al.(2006{\natexlab{b}})Dwork, McSherry, Nissim, and
  Smith]{dwork2006calibrating}
Dwork, C., McSherry, F., Nissim, K., and Smith, A.
\newblock {Calibrating noise to sensitivity in private data analysis}.
\newblock In \emph{TCC}, 2006{\natexlab{b}}.

\bibitem[Dwork et~al.(2014)Dwork, Roth, et~al.]{dwork2014algorithmic}
Dwork, C., Roth, A., et~al.
\newblock {The algorithmic foundations of differential privacy}.
\newblock \emph{Foundations and Trends in Theoretical Computer Science}, 2014.

\bibitem[Farrand et~al.(2020)Farrand, Mireshghallah, Singh, and
  Trask]{farrand2020neither}
Farrand, T., Mireshghallah, F., Singh, S., and Trask, A.
\newblock {Neither private nor fair: Impact of data imbalance on utility and
  fairness in differential privacy}.
\newblock In \emph{Workshop on Privacy-Preserving Machine Learning in
  Practice}, 2020.

\bibitem[Feldman(2020)]{feldman2020does}
Feldman, V.
\newblock {Does learning require memorization? a short tale about a long tail}.
\newblock In \emph{{STOC}}, 2020.

\bibitem[Frigerio et~al.(2019)Frigerio, de~Oliveira, Gomez, and
  Duverger]{frigerio2019differentially}
Frigerio, L., de~Oliveira, A.~S., Gomez, L., and Duverger, P.
\newblock {Differentially private generative adversarial networks for time
  series, continuous, and discrete open data}.
\newblock In \emph{IFIP SEC}, 2019.

\bibitem[Ghalebikesabi et~al.(2021)Ghalebikesabi, Wilde, Jewson, Doucet,
  Vollmer, and Holmes]{ghalebikesabi2021bias}
Ghalebikesabi, S., Wilde, H., Jewson, J., Doucet, A., Vollmer, S., and Holmes,
  C.
\newblock {Bias Mitigated Learning from Differentially Private Synthetic Data:
  A Cautionary Tale}.
\newblock \emph{arXiv:2108.10934}, 2021.

\bibitem[Goodfellow et~al.(2014)Goodfellow, Pouget-Abadie, Mirza, Xu,
  Warde-Farley, Ozair, Courville, and Bengio]{goodfellow2014generative}
Goodfellow, I., Pouget-Abadie, J., Mirza, M., Xu, B., Warde-Farley, D., Ozair,
  S., Courville, A., and Bengio, Y.
\newblock {Generative adversarial nets}.
\newblock \emph{NeurIPS}, 2014.

\bibitem[Hardt et~al.(2016)Hardt, Price, and Srebro]{hardt2016equality}
Hardt, M., Price, E., and Srebro, N.
\newblock {Equality of opportunity in supervised learning}.
\newblock \emph{NeurIPS}, 2016.

\bibitem[Hayes et~al.(2019)Hayes, Melis, Danezis, and
  De~Cristofaro]{hayes2019logan}
Hayes, J., Melis, L., Danezis, G., and De~Cristofaro, E.
\newblock {Logan: Membership inference attacks against generative models}.
\newblock In \emph{{PoPETs}}, 2019.

\bibitem[Holohan et~al.(2019)Holohan, Braghin, Mac~Aonghusa, and
  Levacher]{holohan2019diffprivlib}
Holohan, N., Braghin, S., Mac~Aonghusa, P., and Levacher, K.
\newblock Diffprivlib: the {IBM} differential privacy library.
\newblock \emph{arXiv:1907.02444}, 2019.

\bibitem[Hong(2020)]{census2020}
Hong, J.
\newblock {Census 2020 +/- 2: Census, Differential Privacy, and the Future of
  Data}.
\newblock
  \url{https://www.hawaiidata.org/news/2020/9/24/census2020-census-differential-privacy-future-of-data},
  2020.

\bibitem[Jagielski et~al.(2019)Jagielski, Kearns, Mao, Oprea, Roth,
  Sharifi-Malvajerdi, and Ullman]{jagielski2019differentially}
Jagielski, M., Kearns, M., Mao, J., Oprea, A., Roth, A., Sharifi-Malvajerdi,
  S., and Ullman, J.
\newblock {Differentially private fair learning}.
\newblock In \emph{ICML}, 2019.

\bibitem[Jordon et~al.(2018)Jordon, Yoon, and Van Der~Schaar]{jordon2018pate}
Jordon, J., Yoon, J., and Van Der~Schaar, M.
\newblock {PATE-GAN: Generating synthetic data with differential privacy
  guarantees}.
\newblock In \emph{ICLR}, 2018.

\bibitem[Kaggle(2013)]{kaggle2013purchases}
Kaggle.
\newblock {Acquire Valued Shoppers Challenge}.
\newblock
  \url{https://www.kaggle.com/c/acquire-valued-shoppers-challenge/data}, 2013.

\bibitem[Koller \& Friedman(2009)Koller and Friedman]{koller2009probabilistic}
Koller, D. and Friedman, N.
\newblock \emph{{Probabilistic graphical models: principles and techniques}}.
\newblock MIT Press, 2009.

\bibitem[Kuppam et~al.(2019)Kuppam, McKenna, Pujol, Hay, Machanavajjhala, and
  Miklau]{kuppam2019fair}
Kuppam, S., McKenna, R., Pujol, D., Hay, M., Machanavajjhala, A., and Miklau,
  G.
\newblock {Fair decision making using privacy-protected data}.
\newblock \emph{arXiv:1905.12744}, 2019.

\bibitem[LeCun et~al.(2010)LeCun, Cortes, and Burges]{lecun2010mnist}
LeCun, Y., Cortes, C., and Burges, C.
\newblock {MNIST handwritten digit database}.
\newblock \emph{ATT Labs}, 2010.

\bibitem[Liu et~al.(2021)Liu, Wen, He, Salem, Zhang, Backes, De~Cristofaro,
  Fritz, and Zhang]{liu2021ml}
Liu, Y., Wen, R., He, X., Salem, A., Zhang, Z., Backes, M., De~Cristofaro, E.,
  Fritz, M., and Zhang, Y.
\newblock {ML-Doctor: Holistic Risk Assessment of Inference Attacks Against
  Machine Learning Models}.
\newblock \emph{arXiv:2102.02551}, 2021.

\bibitem[McKenna et~al.(2021)McKenna, Miklau, and Sheldon]{mckenna2021winning}
McKenna, R., Miklau, G., and Sheldon, D.
\newblock {Winning the NIST Contest: A scalable and general approach to
  differentially private synthetic data}.
\newblock \emph{arXiv:2108.04978}, 2021.

\bibitem[McSherry \& Talwar(2007)McSherry and Talwar]{mcsherry2007mechanism}
McSherry, F. and Talwar, K.
\newblock {Mechanism design via differential privacy}.
\newblock In \emph{FOCS}, 2007.

\bibitem[{NHS England}(2021)]{nhs21ae}
{NHS England}.
\newblock {A\&E Synthetic Data}.
\newblock \url{https://data.england.nhs.uk/dataset/a-e-synthetic-data}, 2021.

\bibitem[{NIST}(2018{\natexlab{a}})]{nist2018differential}
{NIST}.
\newblock {2018 Differential Privacy Synthetic Data Challenge}.
\newblock
  \url{https://www.nist.gov/ctl/pscr/open-innovation-prize-challenges/past-prize-challenges/2018-differential-privacy-synthetic},
  2018{\natexlab{a}}.

\bibitem[{NIST}(2018{\natexlab{b}})]{nist2018the}
{NIST}.
\newblock {2018 The Unlinkable Data Challenge}.
\newblock
  \url{https://www.nist.gov/ctl/pscr/open-innovation-prize-challenges/past-prize-challenges/2018-unlinkable-data-challenge},
  2018{\natexlab{b}}.

\bibitem[Papernot et~al.(2016)Papernot, Abadi, Erlingsson, Goodfellow, and
  Talwar]{papernot2016semi}
Papernot, N., Abadi, M., Erlingsson, U., Goodfellow, I., and Talwar, K.
\newblock {Semi-supervised knowledge transfer for deep learning from private
  training data}.
\newblock \emph{arXiv:1610.05755}, 2016.

\bibitem[Papernot et~al.(2018)Papernot, Song, Mironov, Raghunathan, Talwar, and
  Erlingsson]{papernot2018scalable}
Papernot, N., Song, S., Mironov, I., Raghunathan, A., Talwar, K., and
  Erlingsson, {\'U}.
\newblock {Scalable private learning with pate}.
\newblock \emph{arXiv:1802.08908}, 2018.

\bibitem[Pedregosa et~al.(2011)Pedregosa, Varoquaux, Gramfort, Michel, Thirion,
  Grisel, Blondel, Prettenhofer, Weiss, Dubourg, Vanderplas, Passos,
  Cournapeau, Brucher, Perrot, and Duchesnay]{sklearn2011}
Pedregosa, F., Varoquaux, G., Gramfort, A., Michel, V., Thirion, B., Grisel,
  O., Blondel, M., Prettenhofer, P., Weiss, R., Dubourg, V., Vanderplas, J.,
  Passos, A., Cournapeau, D., Brucher, M., Perrot, M., and Duchesnay, E.
\newblock Scikit-learn: Machine learning in {P}ython.
\newblock \emph{Journal of Machine Learning Research}, 2011.

\bibitem[Pereira et~al.(2021)Pereira, Kshirsagar, Mukherjee, Dodhia, and
  Ferres]{pereira2021analysis}
Pereira, M., Kshirsagar, M., Mukherjee, S., Dodhia, R., and Ferres, J.~L.
\newblock {An Analysis of the Deployment of Models Trained on Private Tabular
  Synthetic Data: Unexpected Surprises}.
\newblock \emph{arXiv:2106.10241}, 2021.

\bibitem[Ping et~al.(2017)Ping, Stoyanovich, and Howe]{ping2017data}
Ping, H., Stoyanovich, J., and Howe, B.
\newblock {DataSynthesizer}.
\newblock \url{https://github.com/DataResponsibly/DataSynthesizer}, 2017.

\bibitem[Shokri et~al.(2017)Shokri, Stronati, Song, and
  Shmatikov]{shokri2017membership}
Shokri, R., Stronati, M., Song, C., and Shmatikov, V.
\newblock {Membership inference attacks against machine learning models}.
\newblock In \emph{IEEE S\&P}, 2017.

\bibitem[Stadler et~al.(2022)Stadler, Oprisanu, and
  Troncoso]{stadler2022synthetic}
Stadler, T., Oprisanu, B., and Troncoso, C.
\newblock {Synthetic Data -- Anonymization Groundhog Day}.
\newblock In \emph{Usenix Security}, 2022.

\bibitem[Suriyakumar et~al.(2021)Suriyakumar, Papernot, Goldenberg, and
  Ghassemi]{suriyakumar2021chasing}
Suriyakumar, V.~M., Papernot, N., Goldenberg, A., and Ghassemi, M.
\newblock {Chasing Your Long Tails: Differentially Private Prediction in Health
  Care Settings}.
\newblock In \emph{ACM FAccT}, 2021.

\bibitem[Tantipongpipat et~al.(2021)Tantipongpipat, Waites, Boob, Siva, and
  Cummings]{tantipongpipat2019differentially}
Tantipongpipat, U., Waites, C., Boob, D., Siva, A., and Cummings, R.
\newblock {Differentially private mixed-type data generation for unsupervised
  learning}.
\newblock 2021.

\bibitem[Thompson \& Warzel(2019)Thompson and Warzel]{thompson19twelve}
Thompson, S.~A. and Warzel, C.
\newblock {The Privacy Project: Twelve Million Phones, One Dataset, Zero
  Privacy}.
\newblock
  \url{https://www.nytimes.com/interactive/2019/12/19/opinion/location-tracking-cell-phone.html},
  2019.

\bibitem[Tran et~al.(2021{\natexlab{a}})Tran, Dinh, Beiter, and
  Fioretto]{tran2021fairness}
Tran, C., Dinh, M.~H., Beiter, K., and Fioretto, F.
\newblock {A Fairness Analysis on Private Aggregation of Teacher Ensembles}.
\newblock \emph{arXiv:2109.08630}, 2021{\natexlab{a}}.

\bibitem[Tran et~al.(2021{\natexlab{b}})Tran, Fioretto, and
  Van~Hentenryck]{tran2021differentially}
Tran, C., Fioretto, F., and Van~Hentenryck, P.
\newblock {Differentially Private and Fair Deep Learning: A Lagrangian Dual
  Approach}.
\newblock \emph{AAAI}, 2021{\natexlab{b}}.

\bibitem[Uniyal et~al.(2021)Uniyal, Naidu, Kotti, Singh, Kenfack,
  Mireshghallah, and Trask]{uniyal2021dp}
Uniyal, A., Naidu, R., Kotti, S., Singh, S., Kenfack, P.~J., Mireshghallah, F.,
  and Trask, A.
\newblock {DP-SGD vs PATE: Which Has Less Disparate Impact on Model Accuracy?}
\newblock \emph{arXiv:2106.12576}, 2021.

\bibitem[{US Census Bureau}(2021)]{dpcensus2021}
{US Census Bureau}.
\newblock {Differential Privacy and the 2020 Census}.
\newblock
  \url{https://www.census.gov/library/fact-sheets/2021/differential-privacy-and-the-2020-census.html},
  2021.

\bibitem[Van Der~Schaar \& Maxfield(2020)Van Der~Schaar and
  Maxfield]{schaar20synthetic}
Van Der~Schaar, M. and Maxfield, N.
\newblock {Synthetic data: Breaking the data logjam in machine learning for
  healthcare}.
\newblock \url{https://tinyurl.com/2hr4atnn}, 2020.

\bibitem[Webster et~al.(2019)Webster, Rabin, Simon, and
  Jurie]{webster2019detecting}
Webster, R., Rabin, J., Simon, L., and Jurie, F.
\newblock {Detecting overfitting of deep generative networks via latent
  recovery}.
\newblock In \emph{IEEE CVPR}, 2019.

\bibitem[Wezerek \& Van~Riper(2020)Wezerek and Van~Riper]{changes2020}
Wezerek, G. and Van~Riper, D.
\newblock {Changes to the Census Could Make Small Towns Disappear}.
\newblock
  \url{https://www.nytimes.com/interactive/2020/02/06/opinion/census-algorithm-privacy.html},
  2020.

\bibitem[Xie et~al.(2018)Xie, Lin, Wang, Wang, and Zhou]{xie2018differentially}
Xie, L., Lin, K., Wang, S., Wang, F., and Zhou, J.
\newblock {Differentially private generative adversarial network}.
\newblock \emph{arXiv:1802.06739}, 2018.

\bibitem[Zhang et~al.(2017)Zhang, Cormode, Procopiuc, Srivastava, and
  Xiao]{zhang2017privbayes}
Zhang, J., Cormode, G., Procopiuc, C.~M., Srivastava, D., and Xiao, X.
\newblock {Privbayes: Private data release via bayesian networks}.
\newblock \emph{ACM Transactions on Database Systems}, 2017.

\bibitem[Zhang et~al.(2021)Zhang, Wang, Li, Honorio, Backes, He, Chen, and
  Zhang]{zhang2021privsyn}
Zhang, Z., Wang, T., Li, N., Honorio, J., Backes, M., He, S., Chen, J., and
  Zhang, Y.
\newblock {Privsyn: Differentially private data synthesis}.
\newblock In \emph{USENIX Security}, 2021.

\end{thebibliography}
\bibliographystyle{icml2022}}

\appendix

\section{Additional Results and Plots}
\subsection{S1: Texas and Recall Plots}
\label{app:s1}

\begin{figure*}[t!]
\begin{minipage}{1\textwidth}
	\centering
	\subfigure[\scriptsize PrivBayes]{\includegraphics[width=0.33\textwidth]{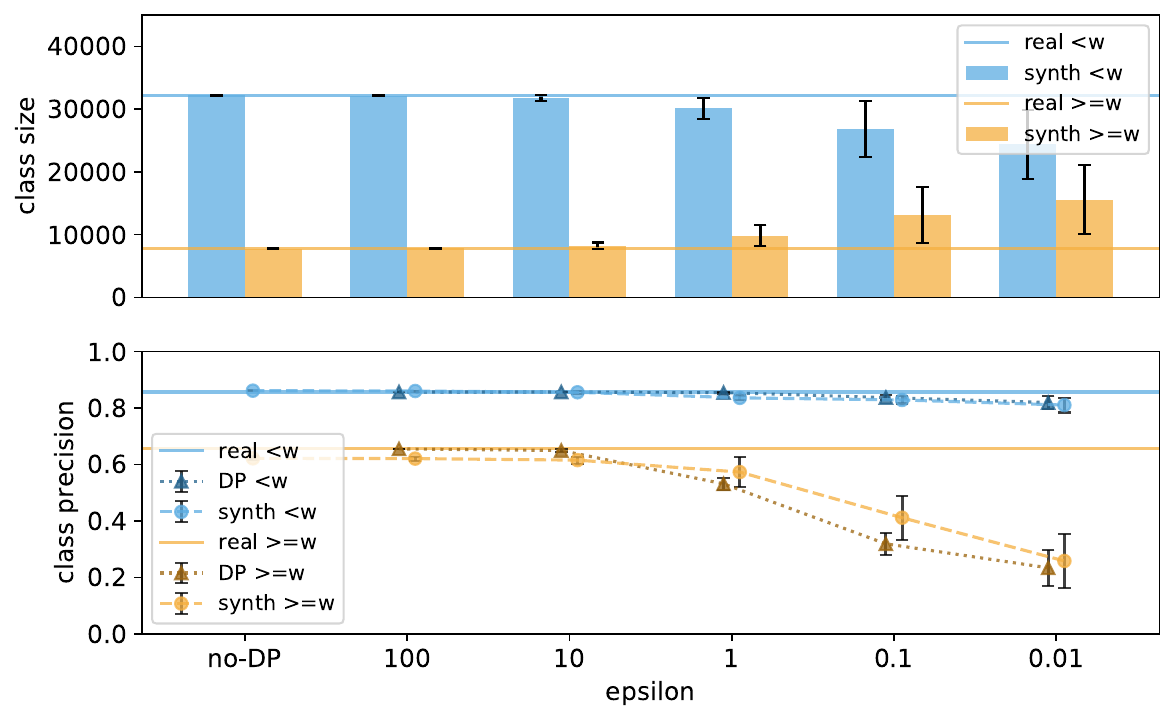}\label{fig:CSPR_T_PrivBayes}}
	\subfigure[\scriptsize DP-WGAN]{\includegraphics[width=0.33\textwidth]{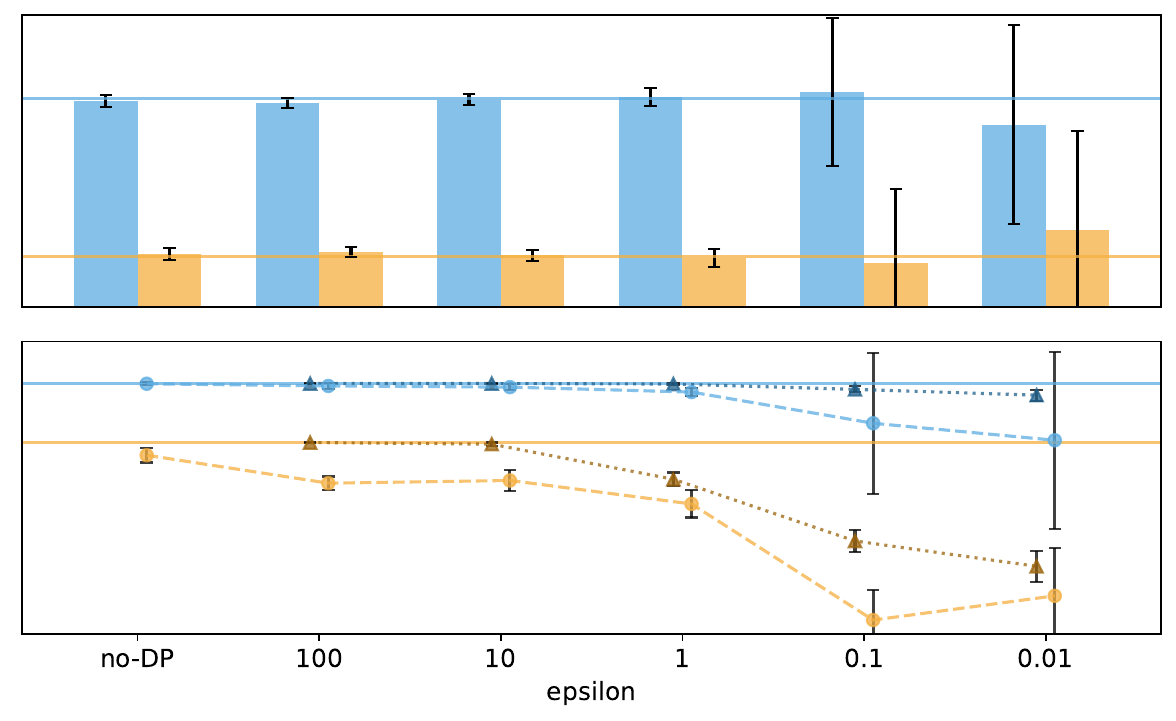}\label{fig:CSPR_T_DPWGAN}}
	\subfigure[\scriptsize PATE-GAN]{\includegraphics[width=0.33\textwidth]{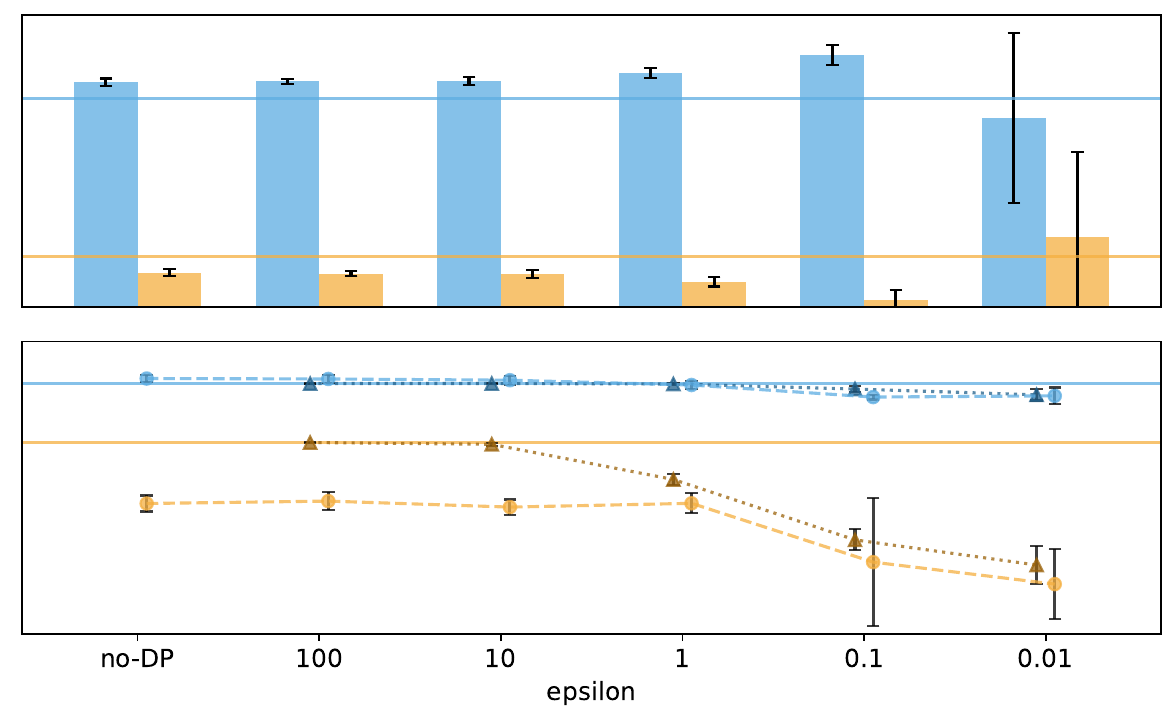}\label{fig:CSPR_T_PATEGAN}}
	\caption{Synthetic data class size (top) and real, DP, and synthetic classifiers precision (bottom) for different levels of $\epsilon$, {\bf\em Texas}, ({\bf\em S1}).}
	\label{fig:CSPR_T}
	\vspace{0.2cm}
	\centering
	\subfigure[\scriptsize PrivBayes]{\includegraphics[width=0.33\textwidth]{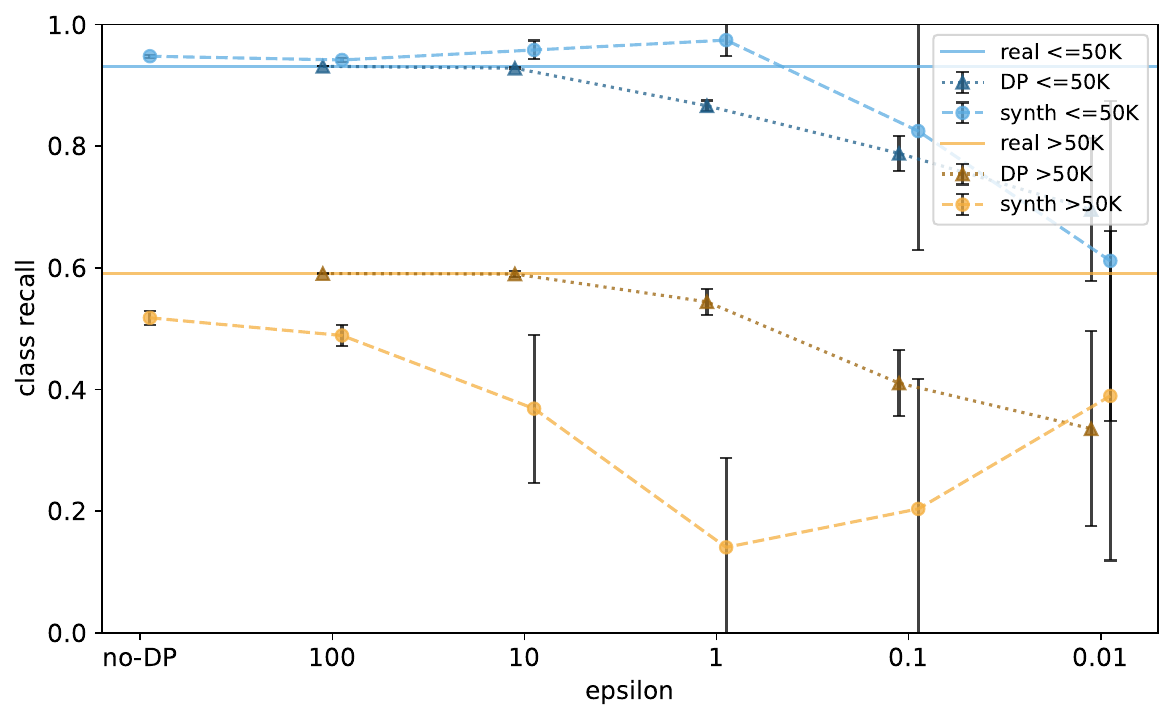}}
	\subfigure[\scriptsize DP-WGAN]{\includegraphics[width=0.33\textwidth]{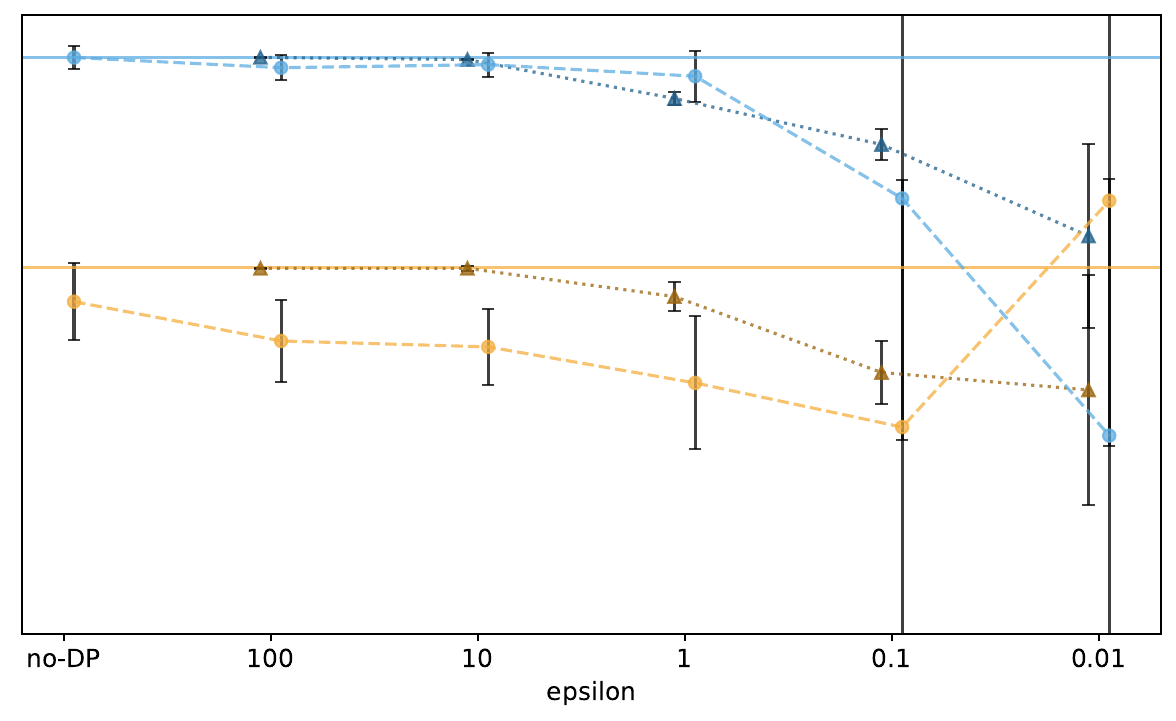}}
	\subfigure[\scriptsize PATE-GAN]{\includegraphics[width=0.33\textwidth]{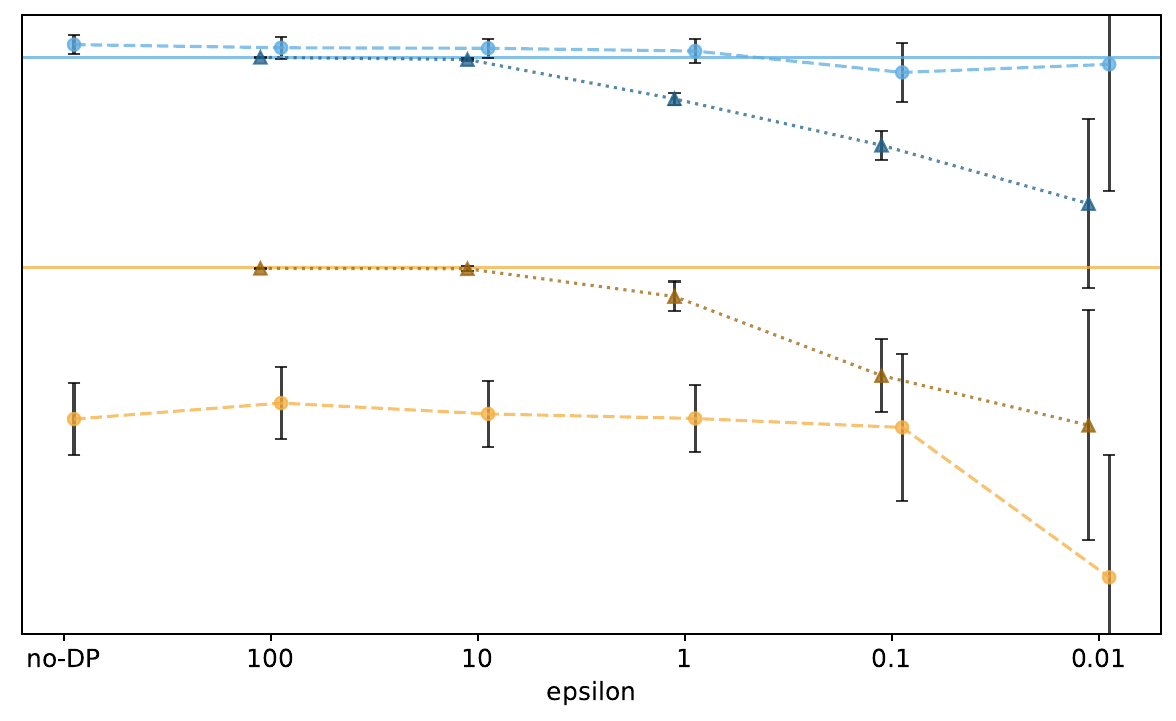}}
	\caption{Real, DP, and synthetic classifiers recall for different levels of $\epsilon$, {\bf\em Adult}, ({\bf\em S1}).}
	\label{fig:aCSPR_A}
		\vspace{0.2cm}
	\centering
	\subfigure[\scriptsize PrivBayes]{\includegraphics[width=0.33\textwidth]{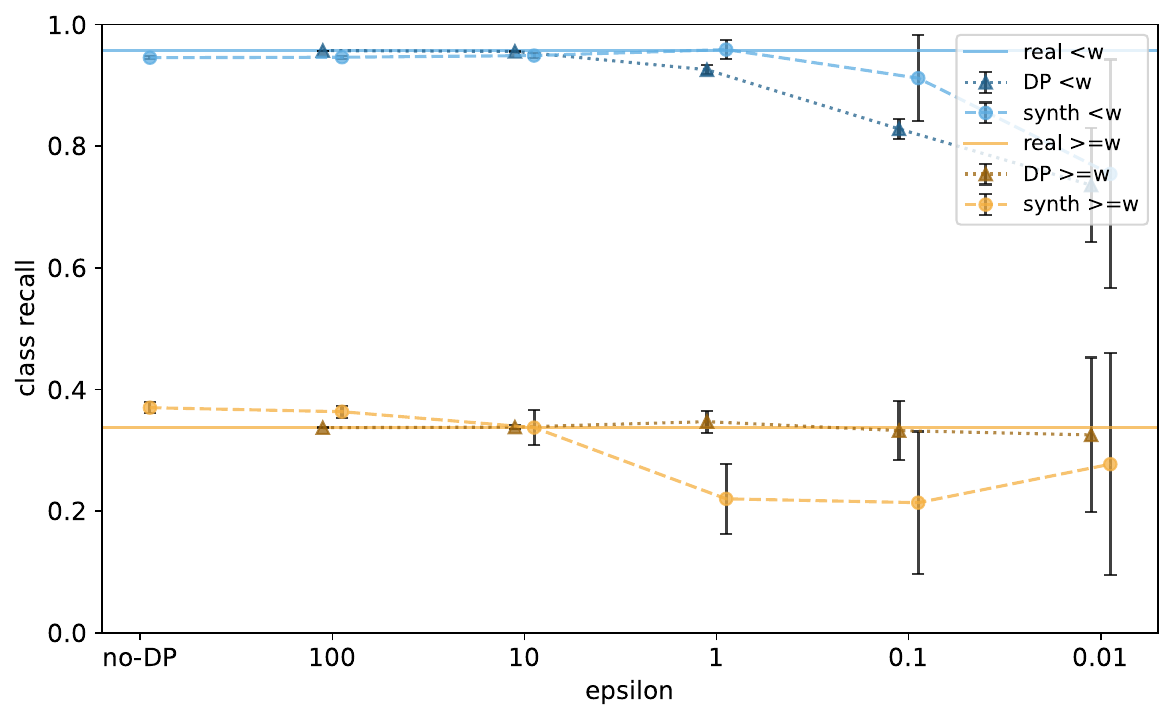}}
	\subfigure[\scriptsize DP-WGAN]{\includegraphics[width=0.33\textwidth]{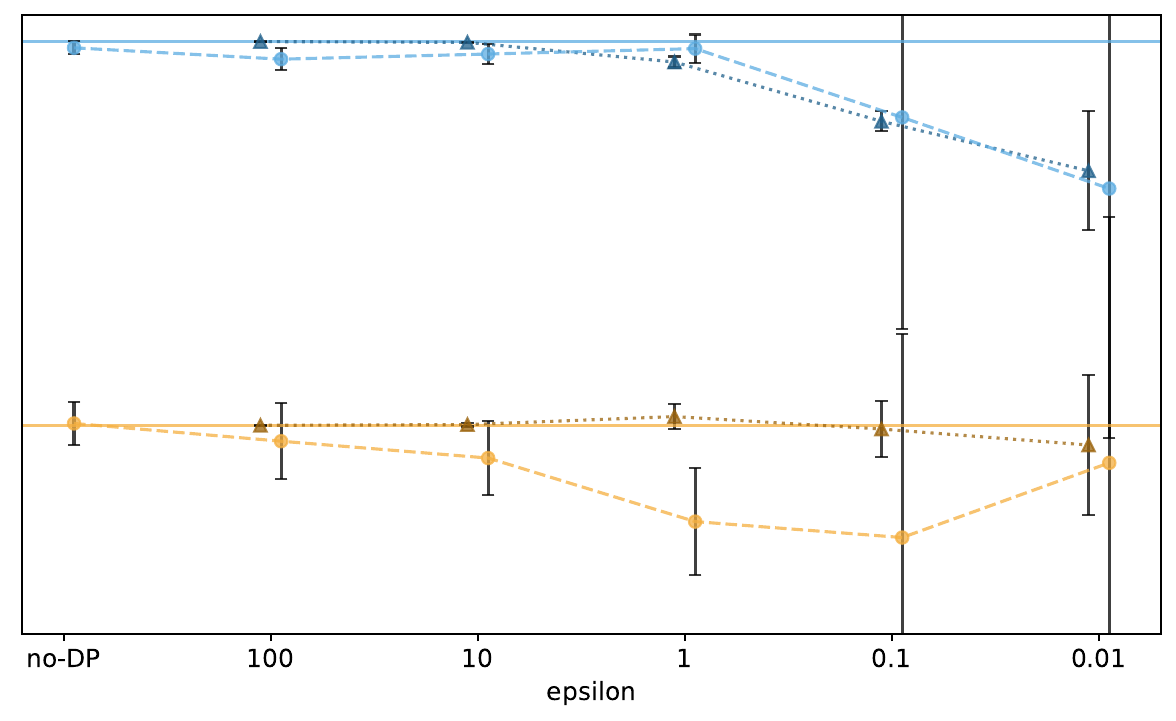}}
	\subfigure[\scriptsize PATE-GAN]{\includegraphics[width=0.33\textwidth]{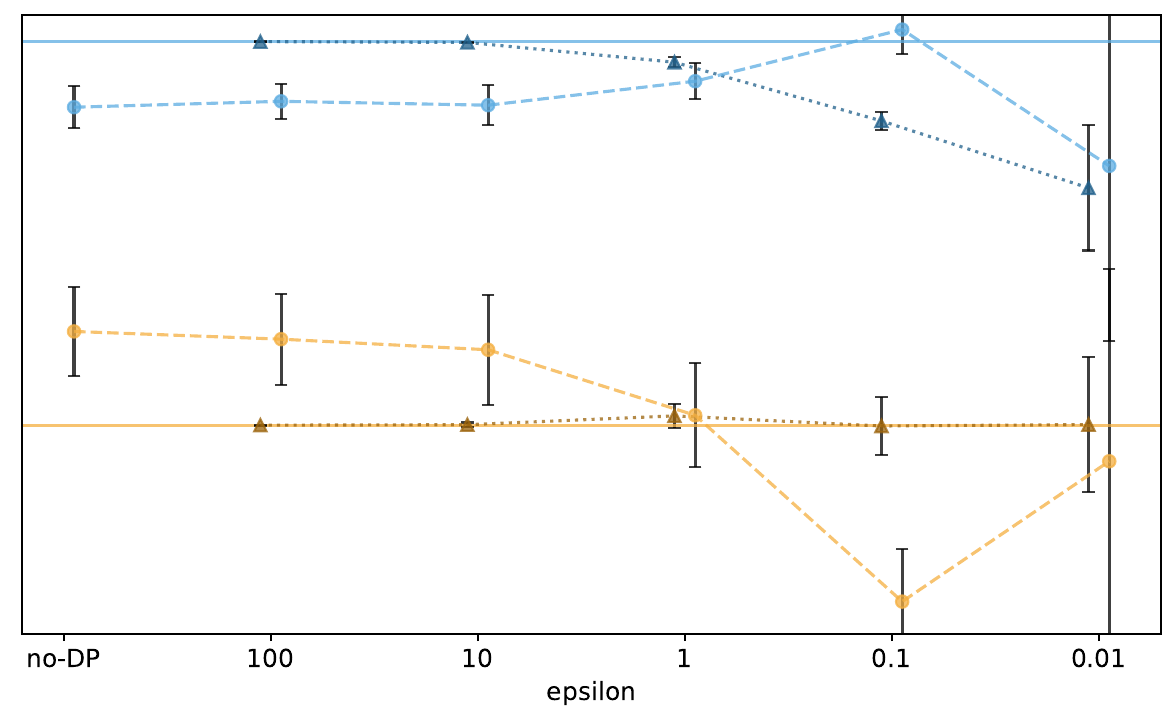}\label{fig:aCSPR_T_PATEGAN}}
	\caption{Real, DP, and synthetic classifiers recall for different levels of $\epsilon$, {\bf\em Texas}, ({\bf\em S1}).}
	\label{fig:aCSPR_T}
		\vspace{0.2cm}
	\centering
	\subfigure[\scriptsize PrivBayes]{\includegraphics[width=0.33\textwidth]{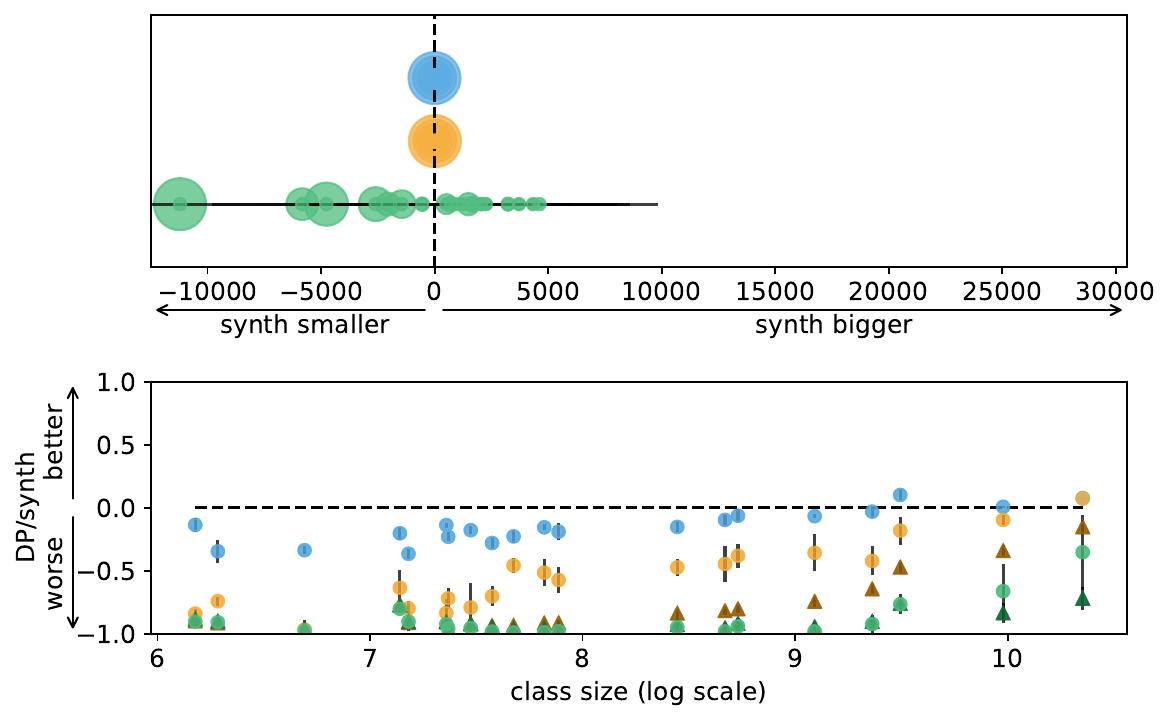}\label{fig:CSPRe_P_PrivBayes}}
	\subfigure[\scriptsize DP-WGAN]{\includegraphics[width=0.33\textwidth]{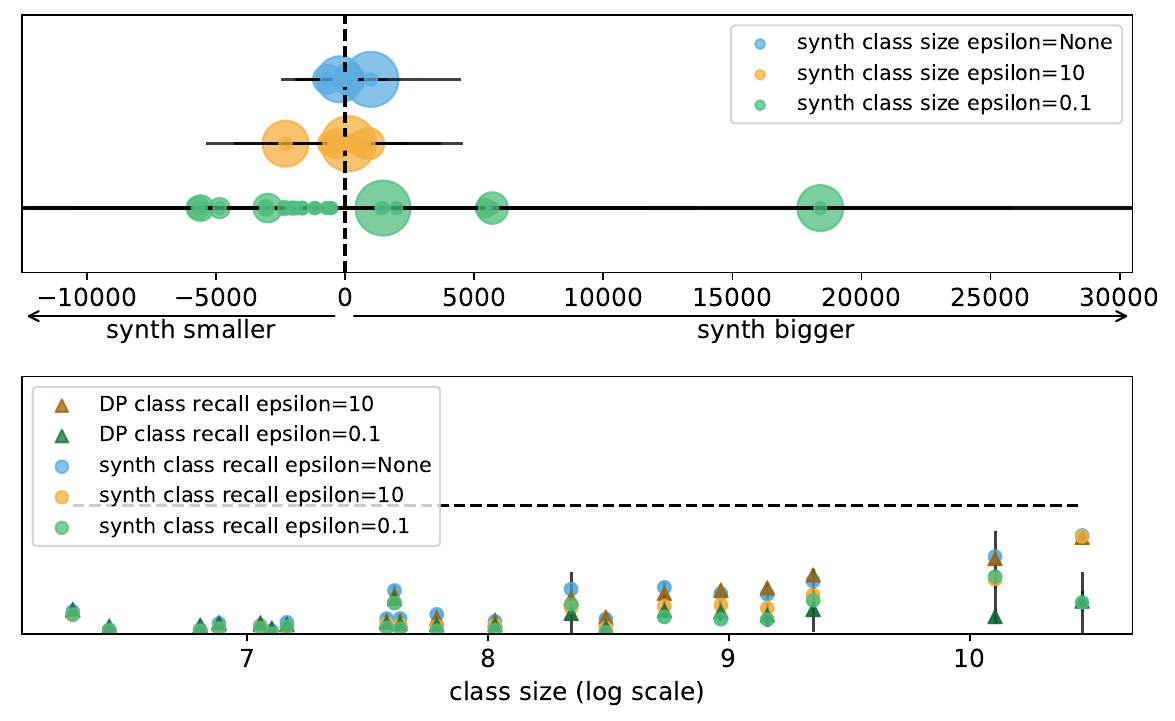}\label{fig:CSPRe_P_DPWGAN}}
 \subfigure[\scriptsize PATE-GAN]{\includegraphics[width=0.33\textwidth]{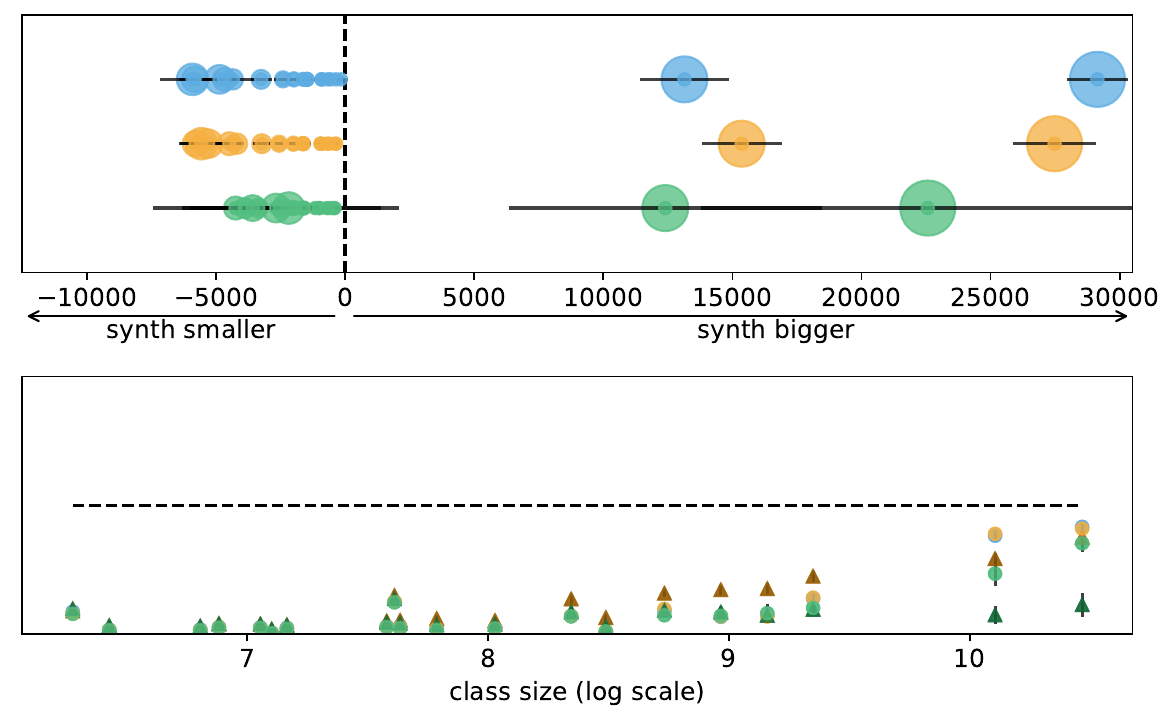}\label{fig:CSPRe_P_PATEGAN}}
	\caption{Synthetic data class (multi-class) size relative to real (top) (each bubble denotes a distinct class while the size its relative count in the real data) and DP and synthetic classifiers recall relative to real (bottom) for different levels of $\epsilon$, {\bf\em Purchases}, ({\bf\em S2}).}
	\label{fig:CSPRe_P}
		\vspace{0.2cm}
	\centering
	\subfigure{\includegraphics[width=0.99\textwidth]{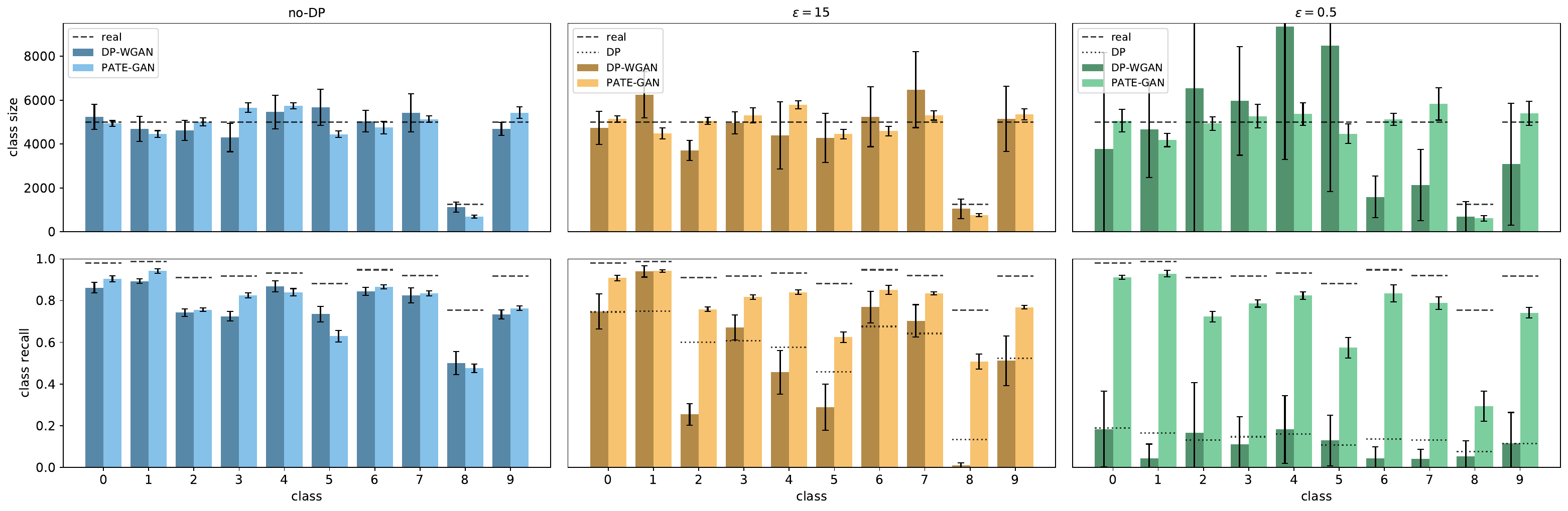}}
	\caption{Synthetic data class (multi-class) size (top) and real, DP, and synthetic classifiers recall (bottom) for different digits and levels of $\epsilon$, {\bf\em MNIST} with class ``8'' downsampled to 0.25 its count, ({\bf\em S2}).}
	\label{fig:CSPR_M_25}
\end{minipage}
\end{figure*}

In Fig.~\ref{fig:CSPR_T} we show the size of the binary class in the real and synthetic data as well as the precision of the real, DP, and synth classifiers for the Texas dataset, they are discussed in Section~\ref{ssec:CSPR}.

In Fig.~\ref{fig:aCSPR_A} and~\ref{fig:aCSPR_T} we plot the recall of the real, DP, and synth classifiers on the Adult and Texas datasets.
For the DP classifiers, recall follows similar patterns as precision for Adult, while, for Texas, there is close to no drop for the underrepresented class and a small drop for the overrepresented class for $\epsilon<0.1$.

For all synth classifiers in the two datasets recall looks more noisy than precision; for most cases (except for PATE-GAN with $\epsilon=0.01$ in Adult), after initially declining with decreasing $\epsilon$ values, the recall of underrepresented class actually starts increasing for $\epsilon<0.1$.
This is most likely because the generated synthetic data is more random; indeed, this is also evident from the large standard deviations in the class size and recall values.
It is interesting to observe that for PATE-GAN in Fig.~\ref{fig:aCSPR_T_PATEGAN}, the underrepresented class recall is actually larger than the real baseline for $\epsilon>1$.

\begin{figure*}[t!]
	\centering
	\subfigure[\scriptsize DP-WGAN, no-DP]{\includegraphics[width=0.495\textwidth]{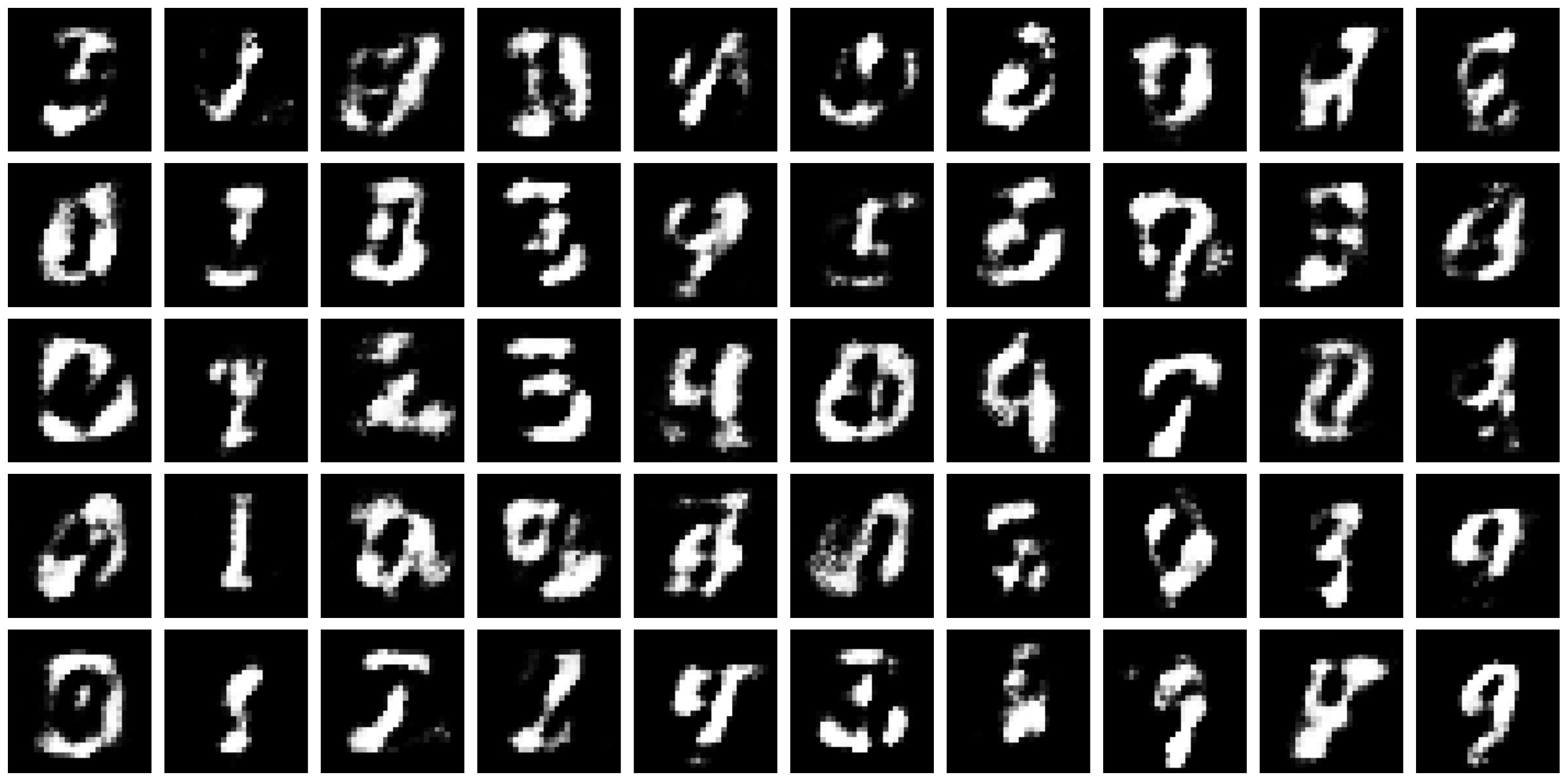}}
	\subfigure[\scriptsize PATE-GAN, no-DP]{\includegraphics[width=0.495\textwidth]{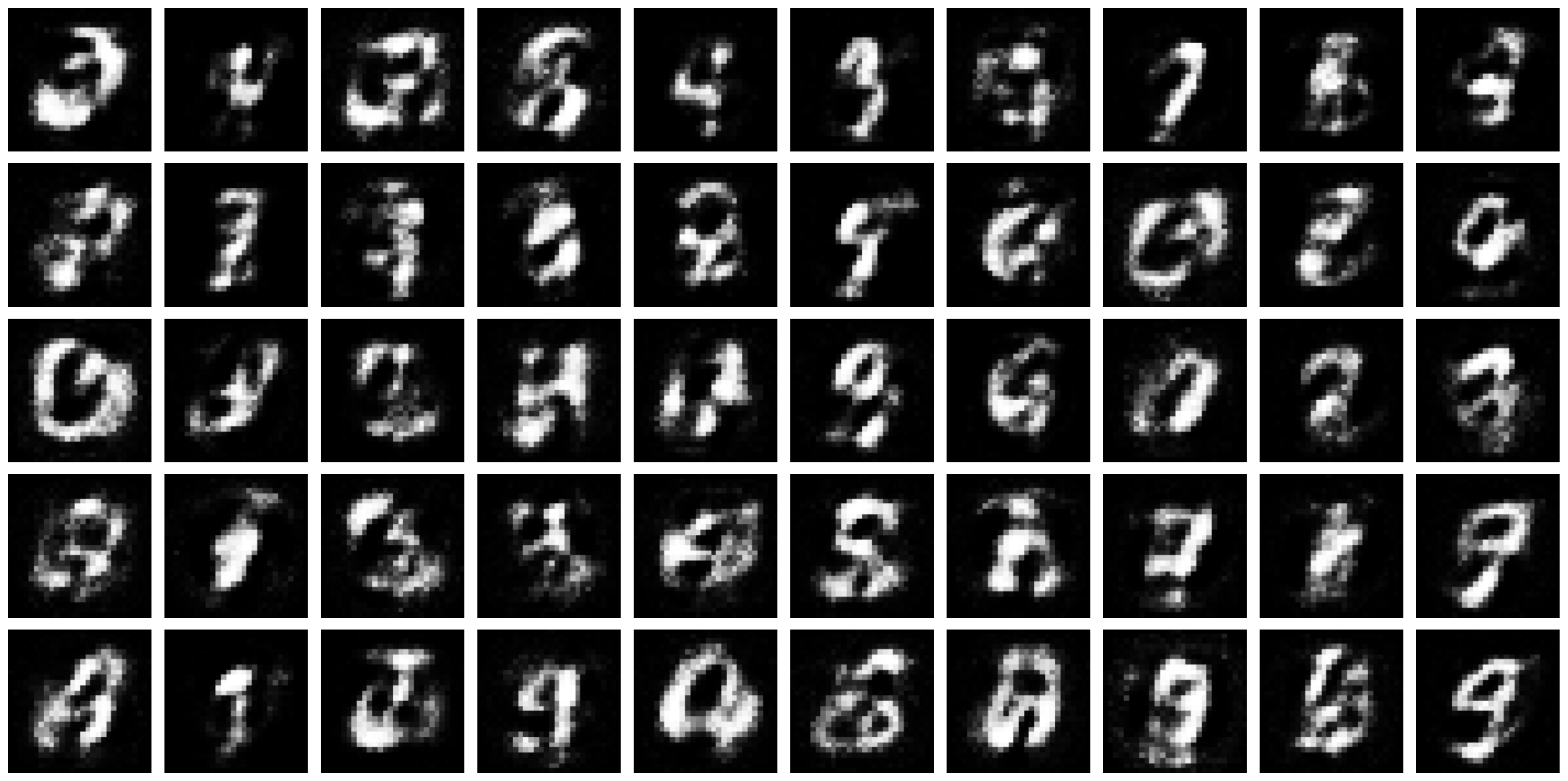}}\\[-1ex]
	\subfigure[\scriptsize DP-WGAN, $\epsilon=15$]{\includegraphics[width=0.495\textwidth]{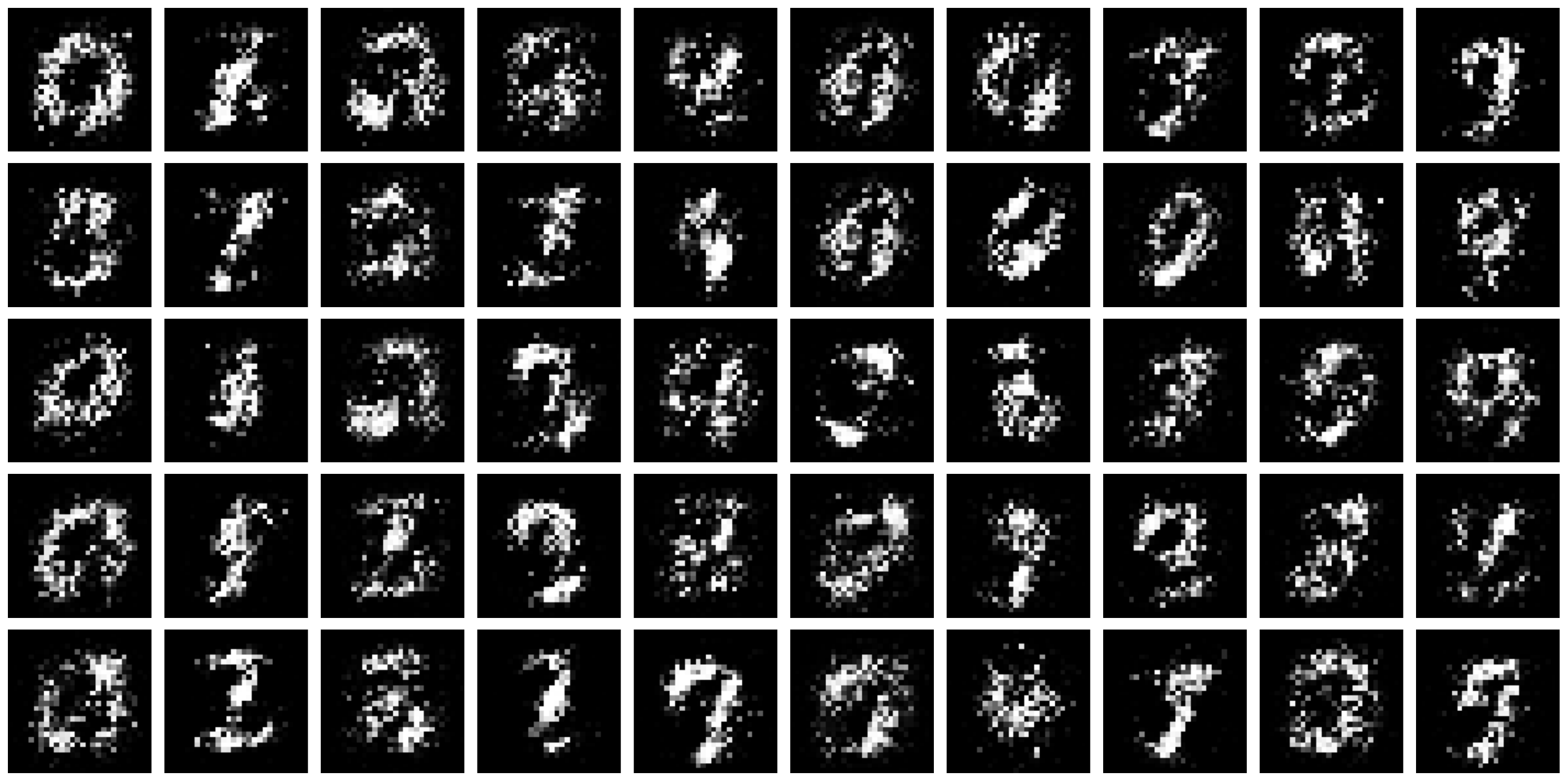}}
	\subfigure[\scriptsize PATE-GAN, $\epsilon=15$]{\includegraphics[width=0.495\textwidth]{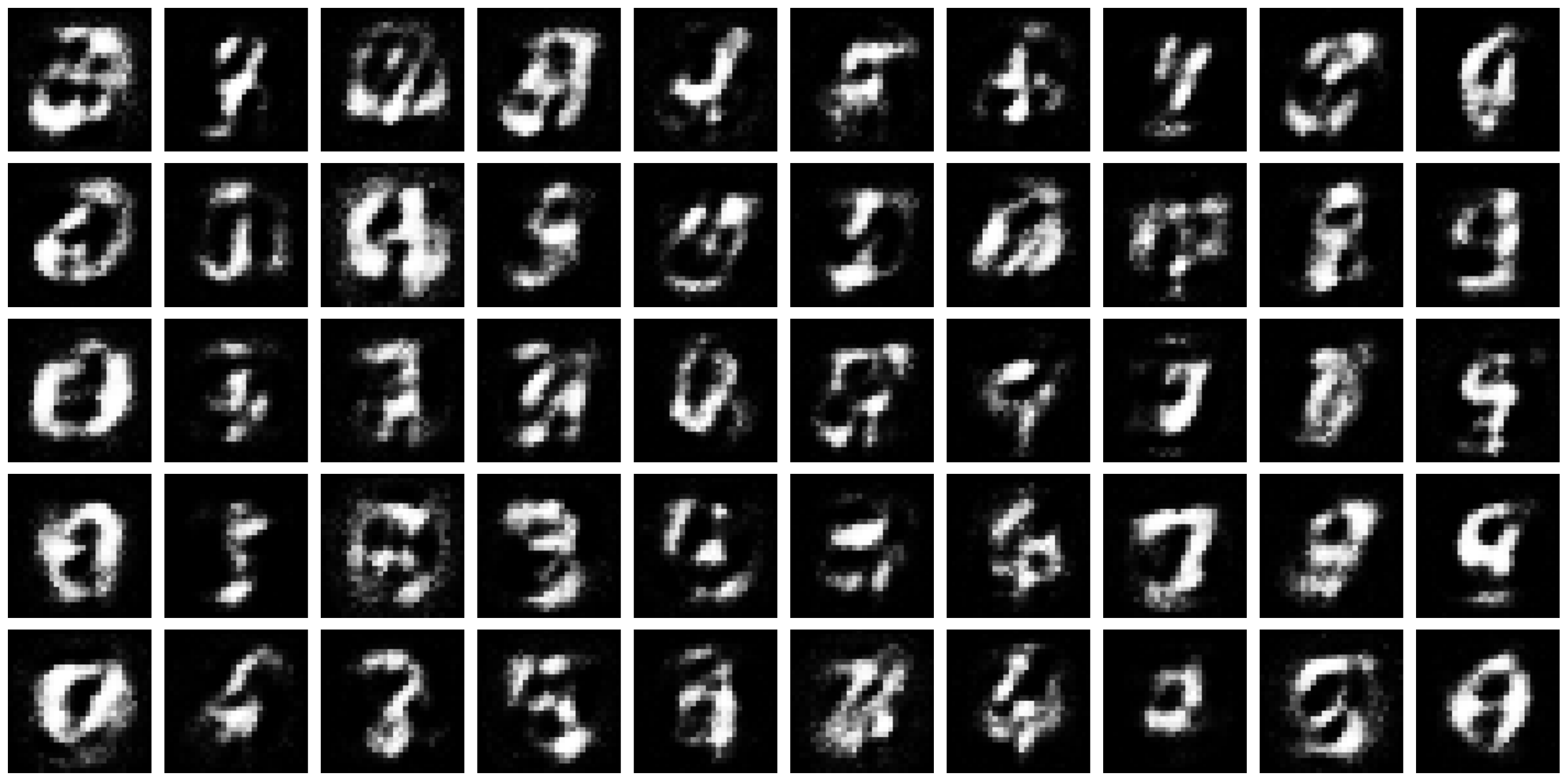}}\\[-1ex]
	\subfigure[\scriptsize DP-WGAN, $\epsilon=5$]{\includegraphics[width=0.495\textwidth]{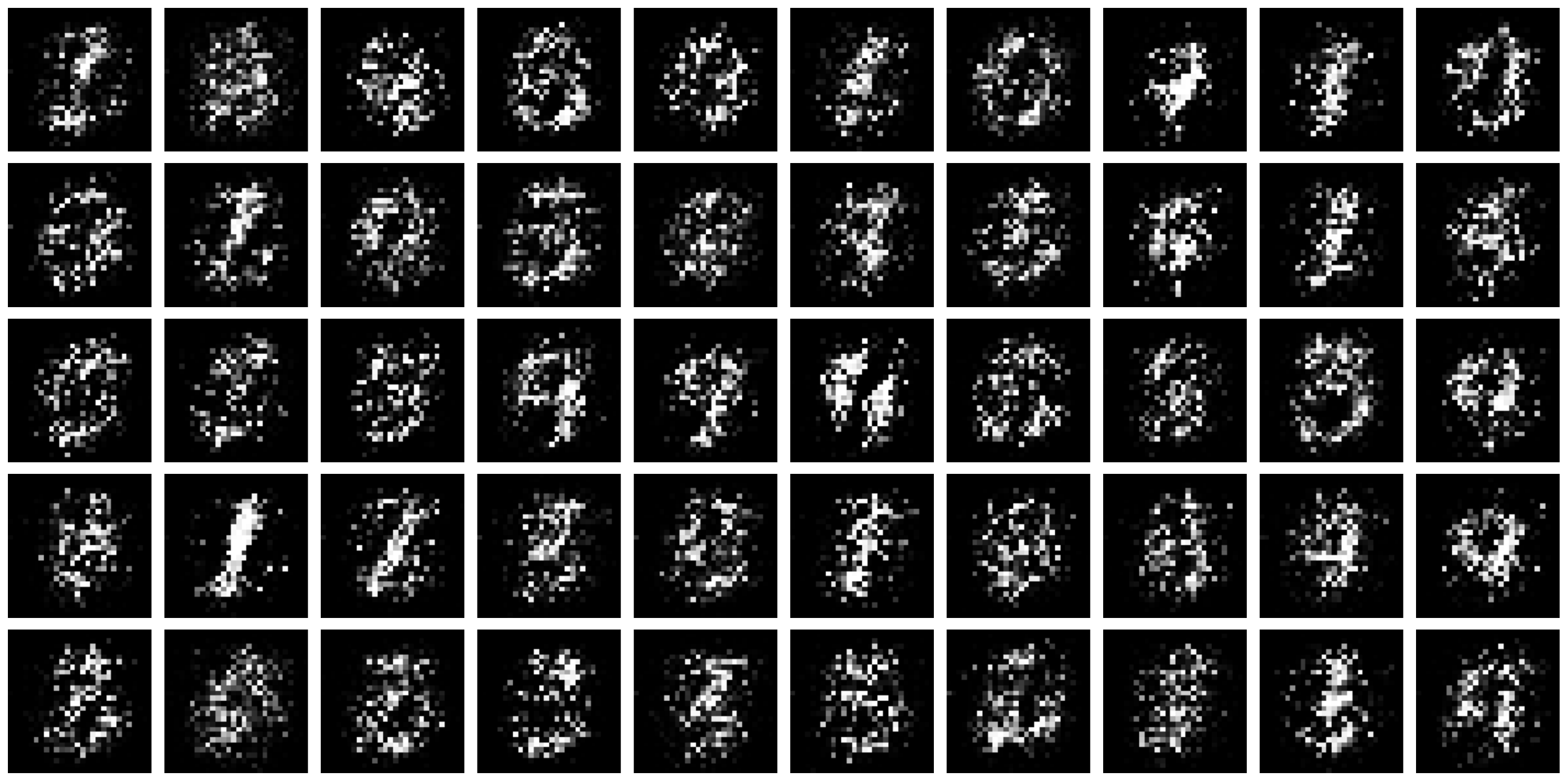}}
	\subfigure[\scriptsize PATE-GAN, $\epsilon=5$]{\includegraphics[width=0.495\textwidth]{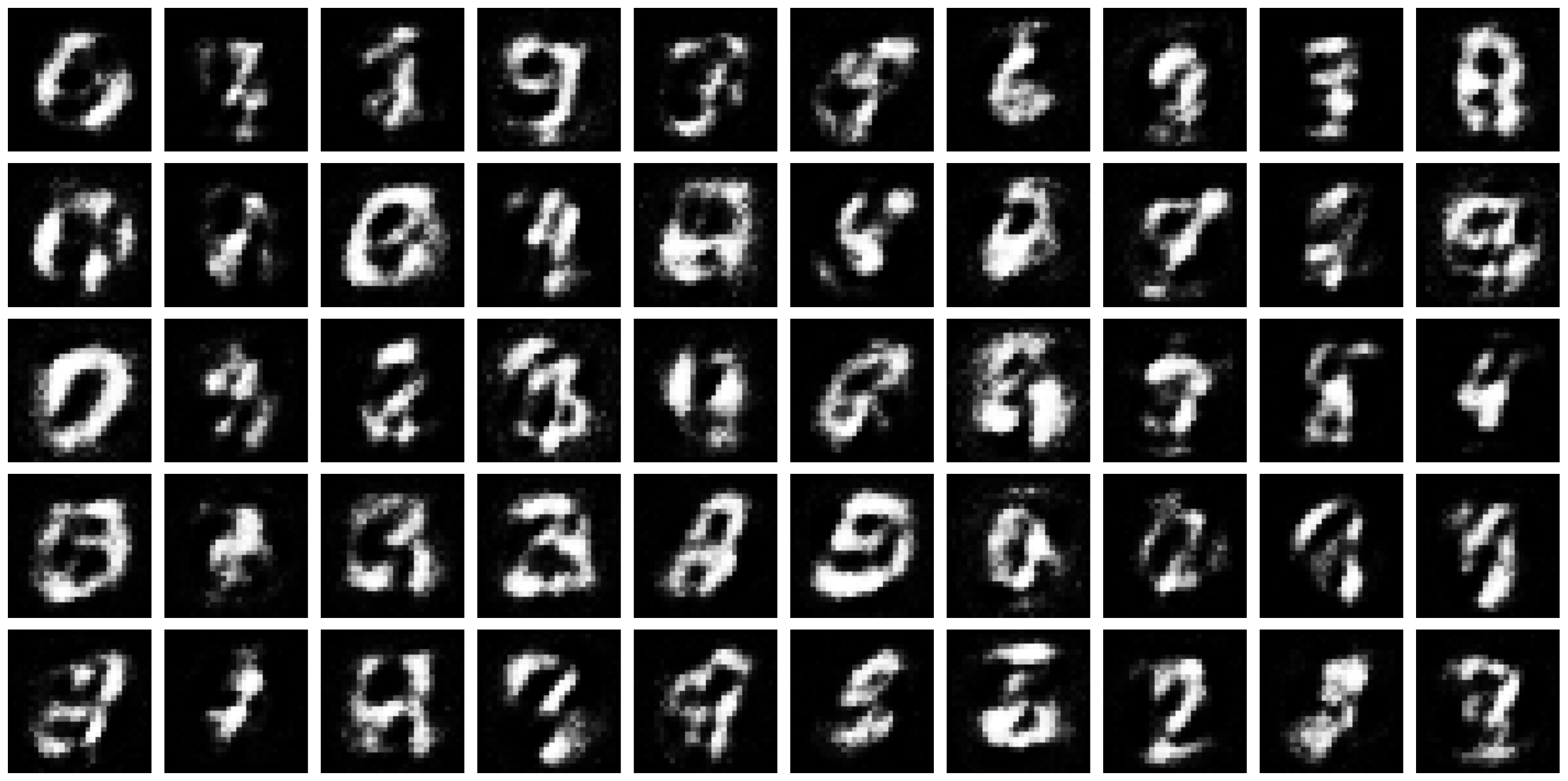}}\\[-1ex]
	\subfigure[\scriptsize DP-WGAN, $\epsilon=0.5$]{\includegraphics[width=0.495\textwidth]{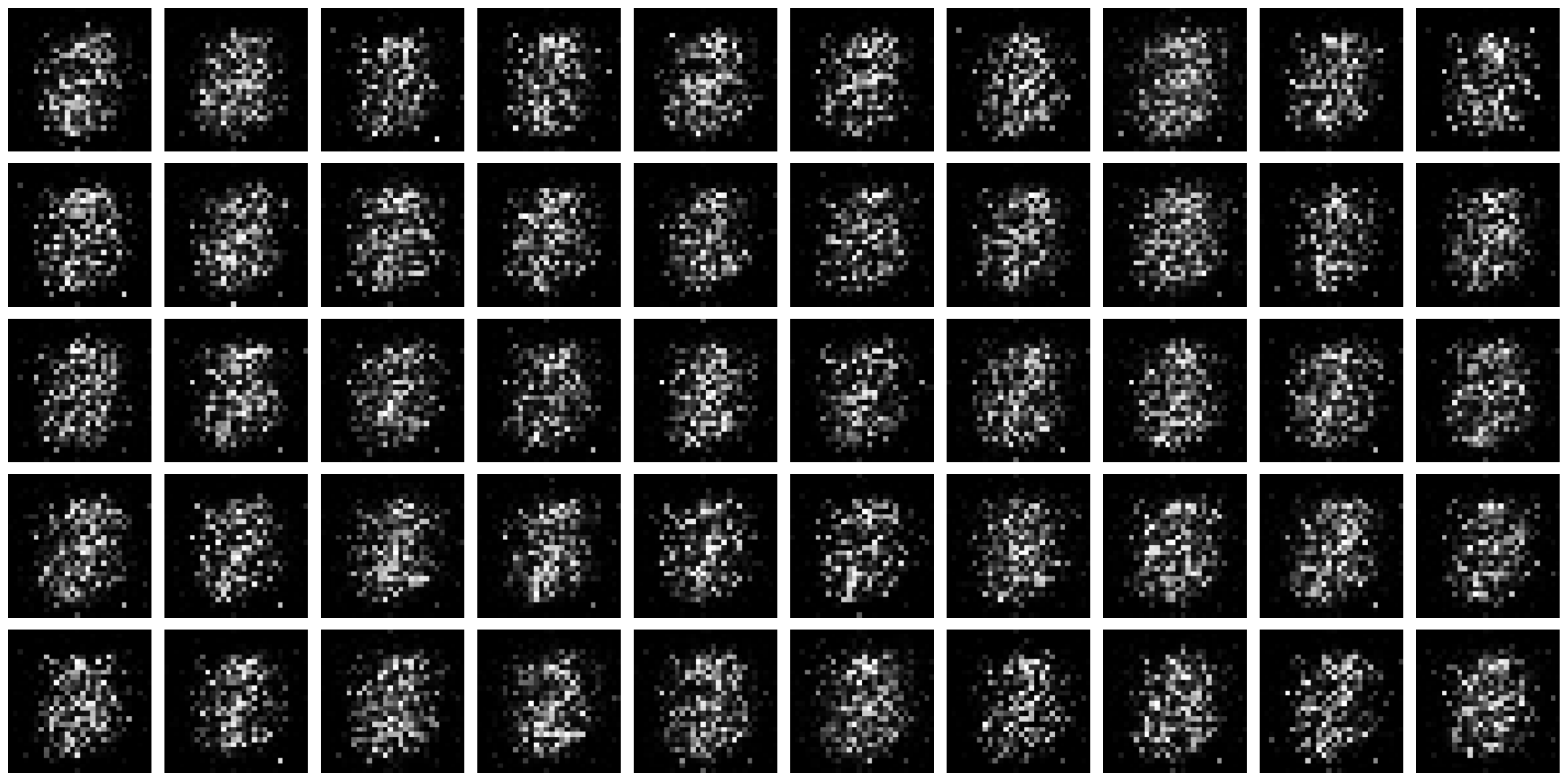}}
	\subfigure[\scriptsize PATE-GAN, $\epsilon=0.5$]{\includegraphics[width=0.495\textwidth]{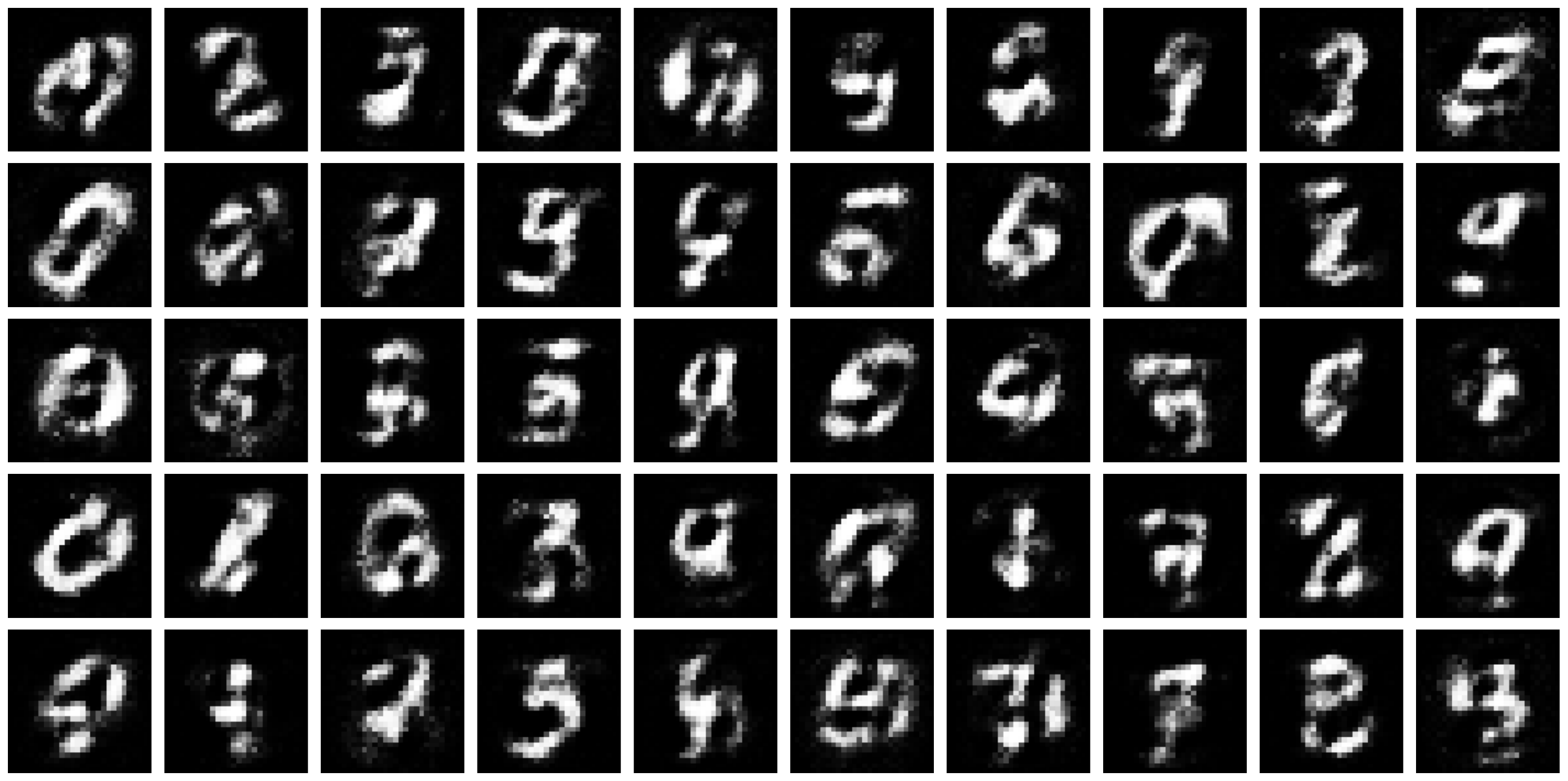}}
	\caption{Synthetic samples (ordered from ``0'' to ``9'' in each subplot) generated by DP-WGAN (left) and PATE-GAN (right) for different $\epsilon$ levels, {\bf\em MNIST} with class ``8'' downsampled to 0.25 its count, ({\bf\em S2}).}
	\label{fig:mnist_digits}
\end{figure*}

\begin{figure*}[t!]
	\centering
	\subfigure[\scriptsize PrivBayes, Adult]{\includegraphics[width=1\textwidth]{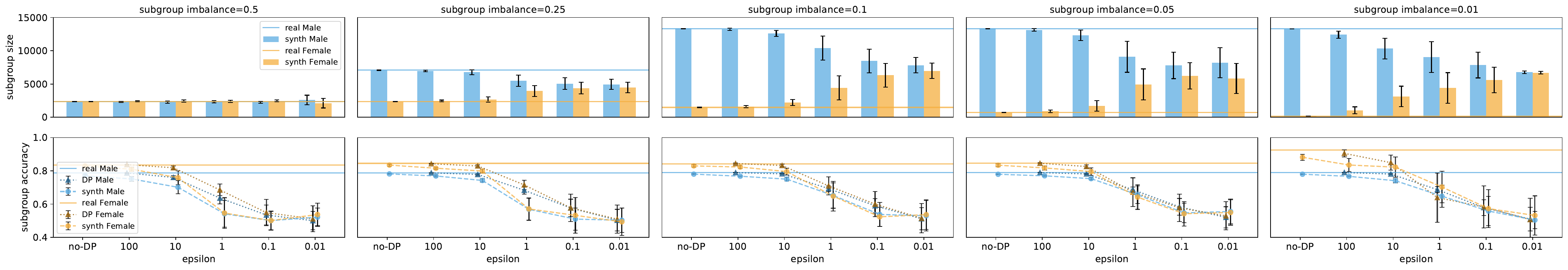}\label{fig:SSA_A_PrivBayes}}
	\subfigure[\scriptsize DP-WGAN, Adult]{\includegraphics[width=1\textwidth]{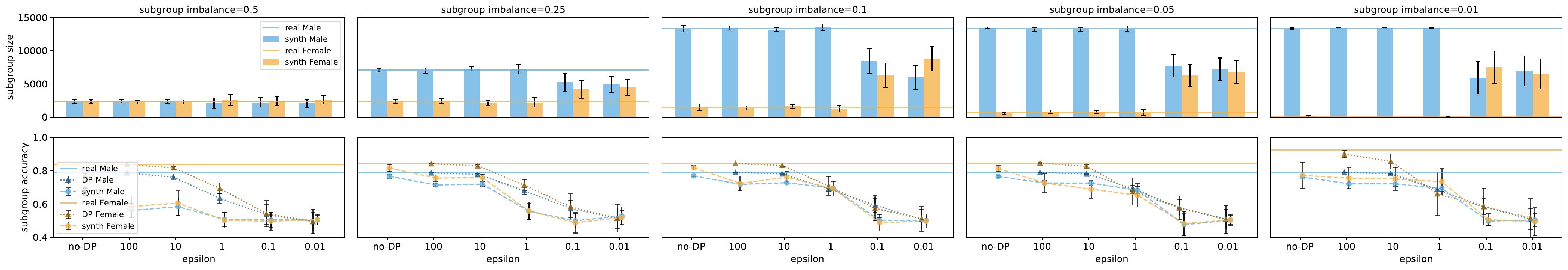}\label{fig:SSA_A_DPWGAN}}
	\subfigure[\scriptsize PATE-GAN, Adult]{\includegraphics[width=1\textwidth]{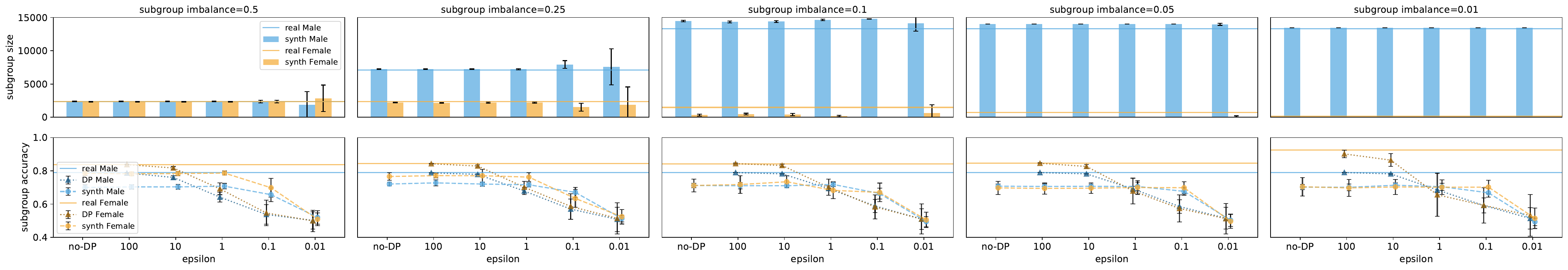}\label{fig:SSA_A_PATEGAN}}
	\subfigure[\scriptsize PrivBayes, Texas]{\includegraphics[width=1\textwidth]{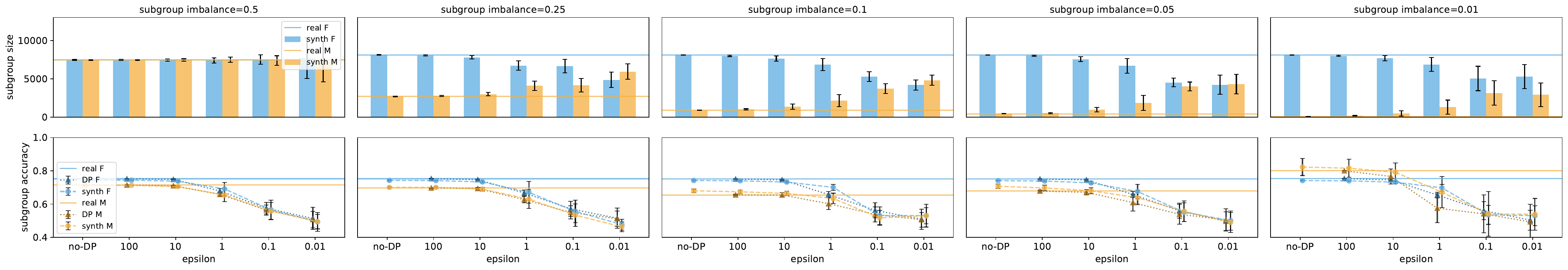}\label{fig:SSA_T_PrivBayes}}
	\subfigure[\scriptsize DP-WGAN, Texas]{\includegraphics[width=1\textwidth]{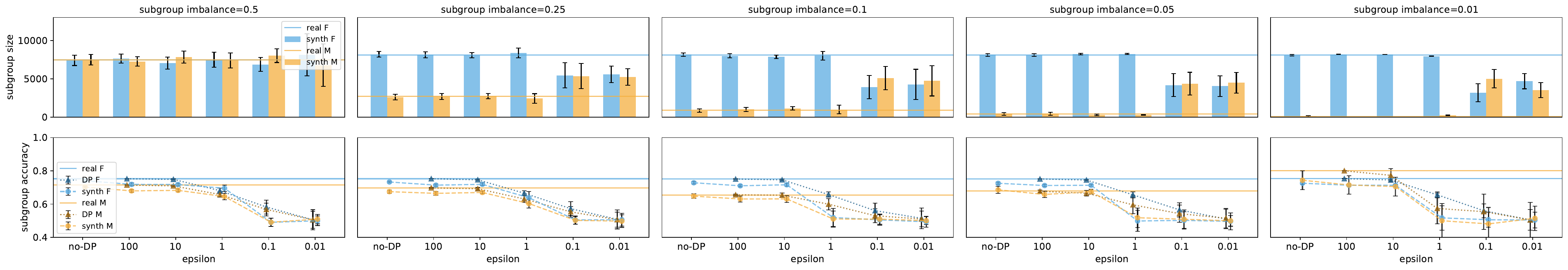}\label{fig:SSA_T_DPWGAN}}
	\subfigure[\scriptsize PATE-GAN, Texas]{\includegraphics[width=1\textwidth]{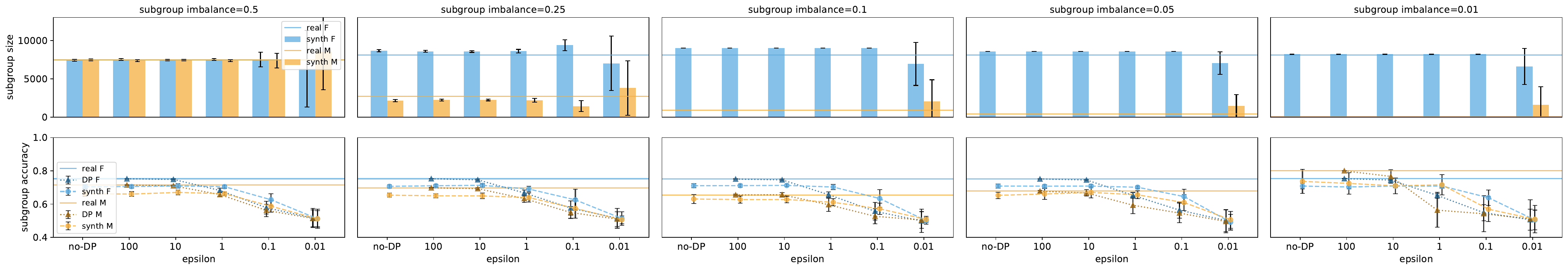}\label{fig:SSA_T_PATEGAN}}
	\caption{Synthetic data single-attribute (sex) subgroup size (top) and real, DP, and synthetic classifiers recall accuracy (bottom) for different single-attribute subgroup imbalance and $\epsilon$ levels, {\bf\em Adult} (top 3) and {\bf\em Texas} (bottom 3), ({\bf\em S3}).}
	\label{fig:SSA}
\end{figure*}

\begin{figure*}[t!]
	\centering
	\subfigure[\scriptsize PrivBayes, Adult]{\includegraphics[width=1\textwidth]{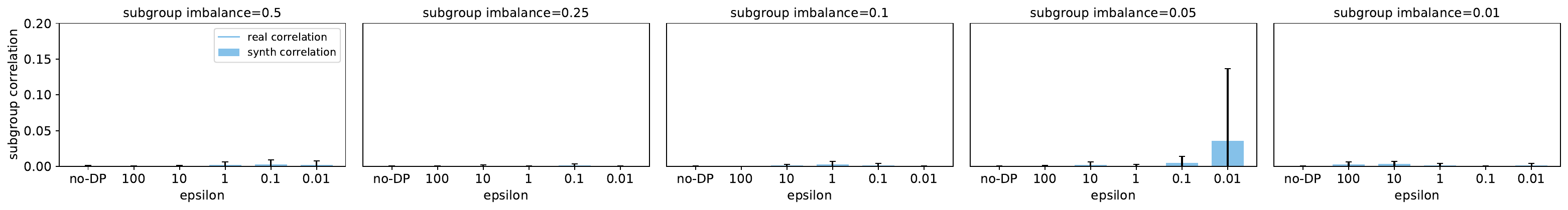}}
	\subfigure[\scriptsize DP-WGAN, Adult]{\includegraphics[width=1\textwidth]{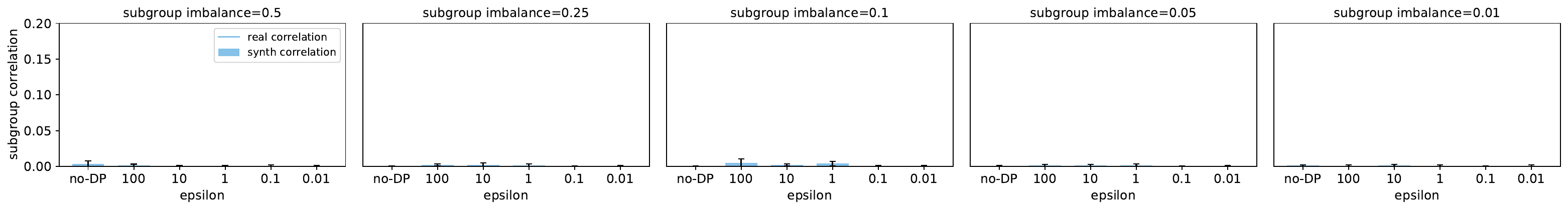}}
	\subfigure[\scriptsize PATE-GAN, Adult]{\includegraphics[width=1\textwidth]{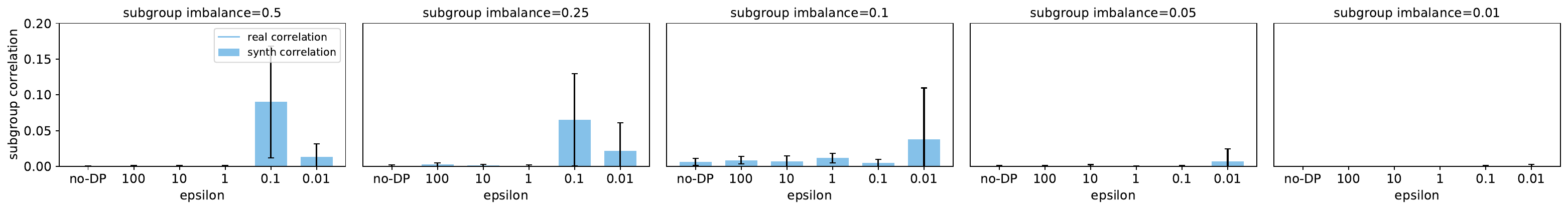}}
	\subfigure[\scriptsize PrivBayes, Texas]{\includegraphics[width=1\textwidth]{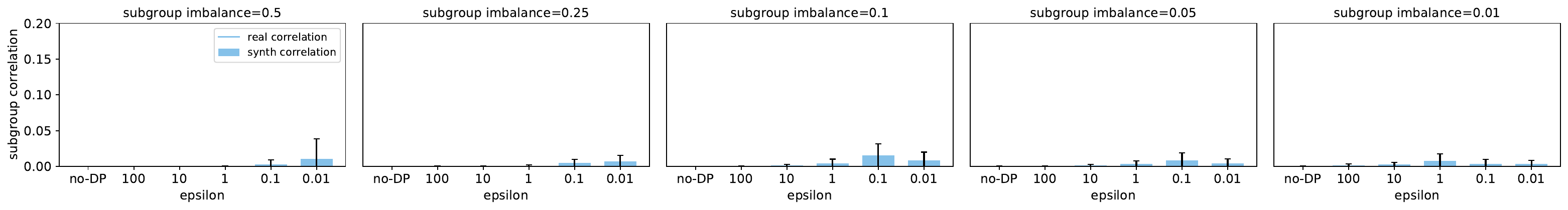}}
	\subfigure[\scriptsize DP-WGAN, Texas]{\includegraphics[width=1\textwidth]{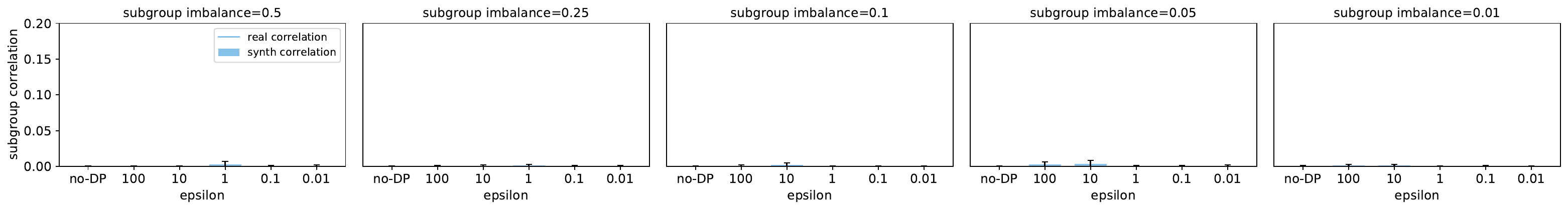}}
	\subfigure[\scriptsize PATE-GAN, Texas]{\includegraphics[width=1\textwidth]{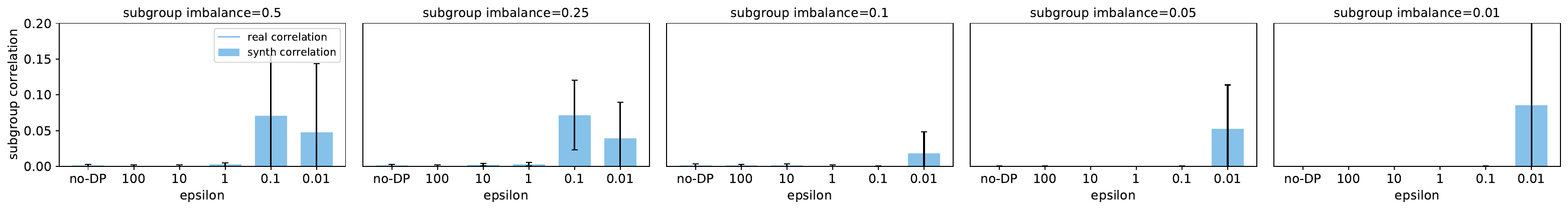}}
	\caption{Mutual information between the single-attribute subgroup (sex) and the target (income/length of stay) columns for different single-attribute subgroup imbalance and $\epsilon$ levels, {\bf\em Adult} (top 3) and {\bf\em Texas} (bottom 3), ({\bf\em S3}).}
	\label{fig:sS_MI}
\end{figure*}

\subsection{S2: Further Purchases and MNIST Plots}
\label{app:mnist}

For the Purchases dataset, Fig.~\ref{fig:CSPRe_P} is duplicate of Fig.~\ref{fig:CSPR_P} but with included error bars.
Looking at the top row, where we display the size of the classes, we observe that PrivBayes has lowest standard deviation (approximately none for ``no-DP'' and $\epsilon=10$), while DP-WGAN the highest.
For PATE-GAN, unlike the other two models, bigger classes exhibit larger standard deviation.
Finally, observing the recall in the bottom row, PATE-GAN classifiers have the lowest variation.

For MNIST, Fig.~\ref{fig:CSPR_M_25} displays the size and recall on all digits for MNIST with class ``8'' undersampled to 0.25 its original size.
In Fig.~\ref{fig:mnist_digits} we can see random synthetic samples produced by DP-WGAN and PATE-GAN for various $\epsilon$ budgets.
The results from both plots are analyzed in Section~\ref{ssec:mCSPR}.
We do not plot or analyze the precision for S2, as the trends are almost identical to recall.

\subsection{S3: Full Plots}
\label{app:s3}

Due to space limitation, in Fig.~\ref{fig:SSA} and~\ref{fig:sS_MI} we plot the full set of experiments that are discussed in detail in Section~\ref{ssec:SSA}.

\begin{figure*}[t!]
	\centering
	\subfigure{\includegraphics[width=0.495\textwidth]{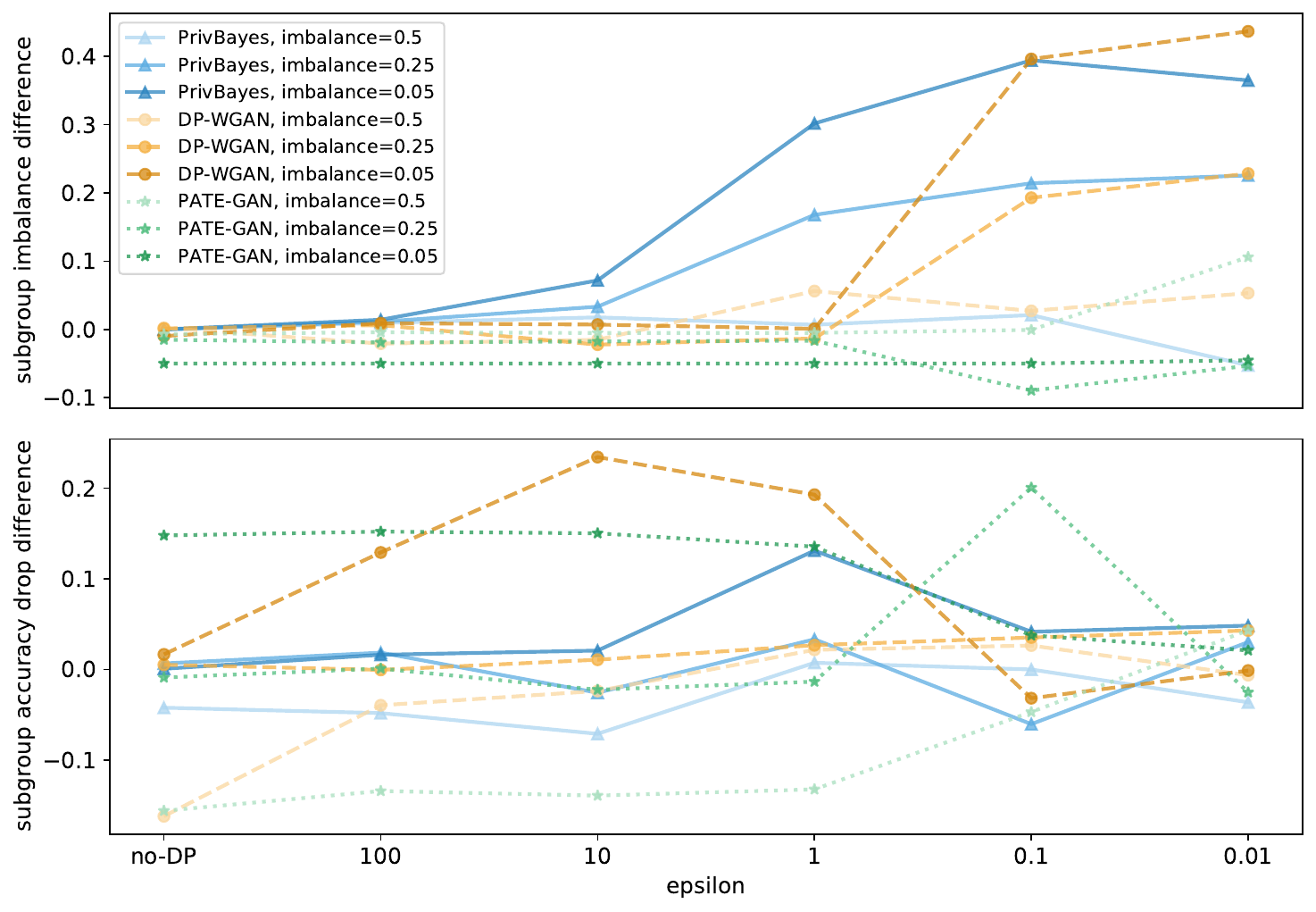}}
	\caption{Minority single-attribute (sex) subgroup imbalance level difference (top) and minority subgroup accuracy drop difference (bottom) relative to majority for different subgroup imbalance and $\epsilon$ levels, {\bf\em Adult}, ({\bf\em S3}).}
	\label{fig:s3_A}
\end{figure*}

Additionally, in Fig.~\ref{fig:s3_A} we plot a summary of the overall trends for the Adult dataset.
From the top row, it could be seen that if there is imbalance in the subgroups PrivBayes balances the data set (dark blue lines have positive slope), DP-WGAN maintains well the imbalance for $\epsilon<1$ (orange dashed lines stay around 0), while PATE-GAN increases it (dotted green lines are negative).
Looking at the bottom row, it could be seen that almost all lines have a positive slope, meaning that with decreased $\epsilon$ the accuracy of the minority subgroup drop quicker than the majority.
Furthermore, darker lines (i.e., more imbalanced datasets) tend to be on top of lighter (i.e., more balanced datasets) which means that increasing the imbalance results in even larger/quicker minority subgroup accuracy drop relative to the majority.

\subsection{S4: Texas Plots}
\label{app:s4texas}

\begin{figure*}[t!]
	\centering
	\subfigure[\scriptsize PrivBayes]{\includegraphics[width=0.65\textwidth]{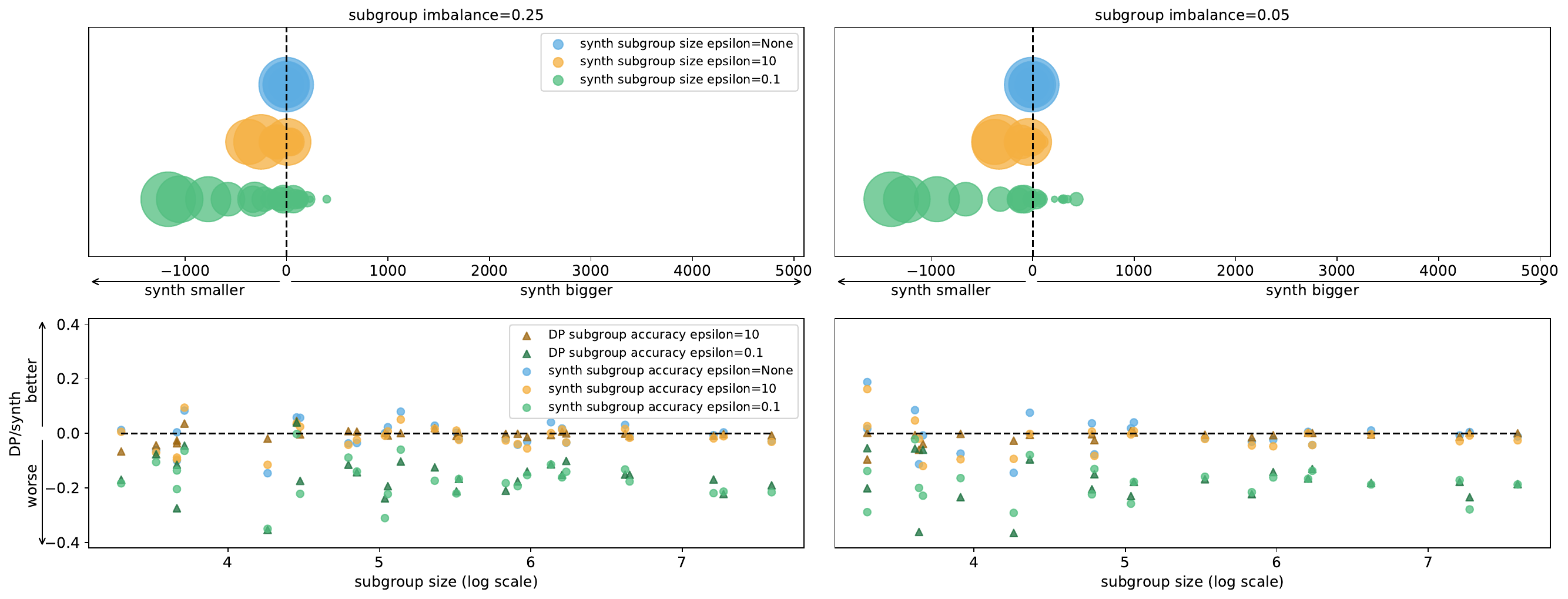}\label{fig:SsSA_T_PrivBayes}}
	\subfigure[\scriptsize DP-WGAN]{\includegraphics[width=0.65\textwidth]{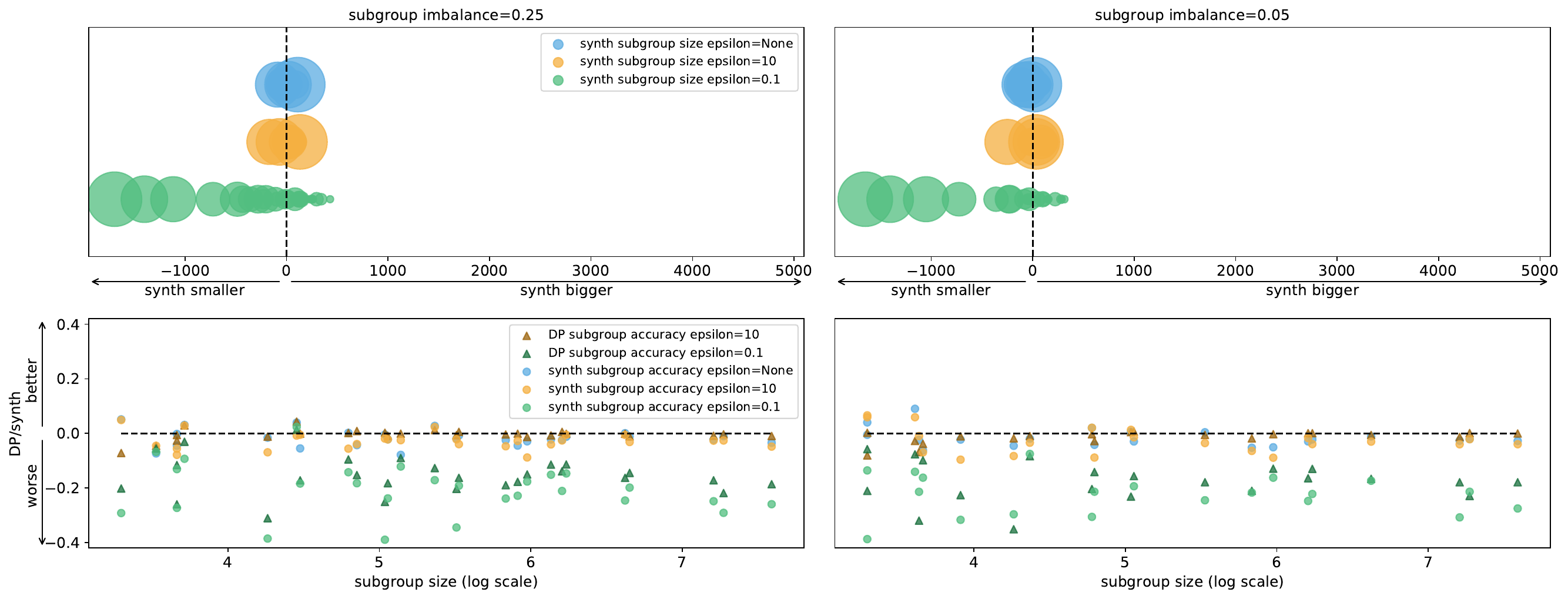}\label{fig:SsSA_T_DPWGAN}}
	\subfigure[\scriptsize PATE-GAN]{\includegraphics[width=0.65\textwidth]{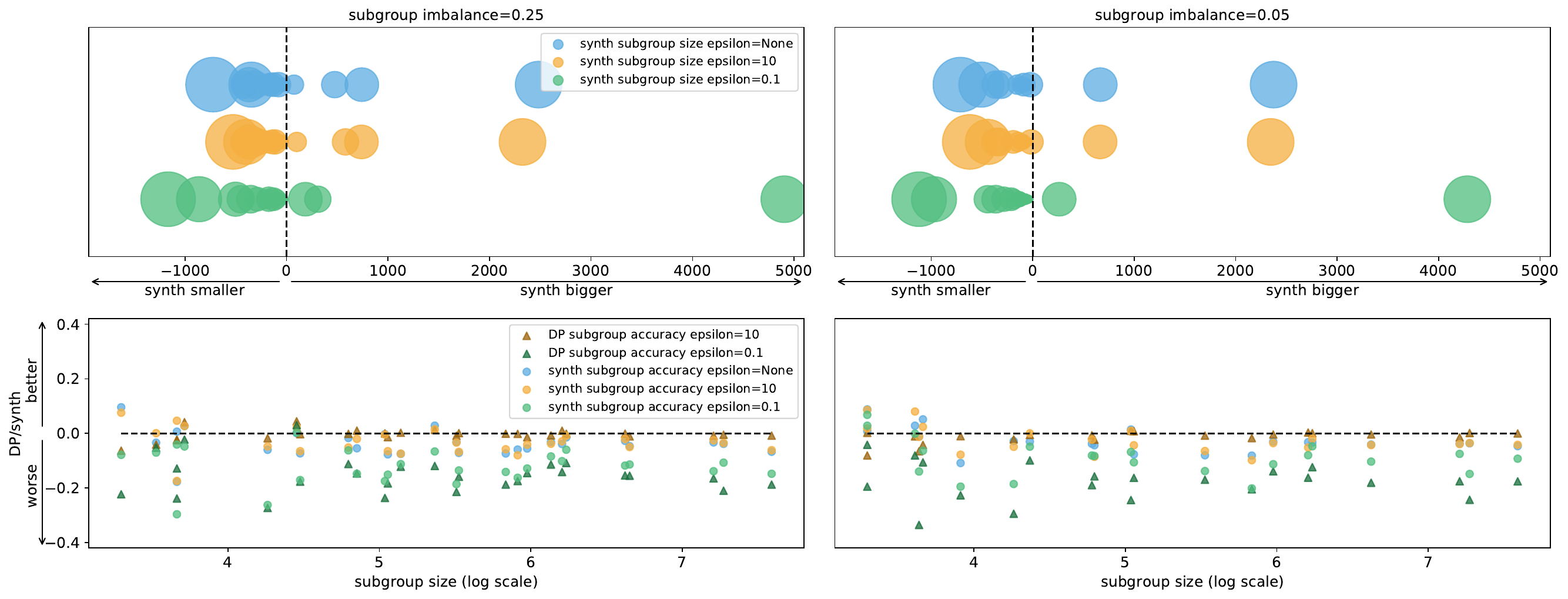}\label{fig:SsSA_T_PATEGAN}}
	\caption{Synthetic data multi-attribute (intersection of age, sex, and race) subgroup size relative to real (top) (each bubble denotes a distinct subgroup while the size its relative count in the real data) and DP and synthetic classifiers accuracy relative to real (bottom) for different single-attribute (sex) subgroup imbalance and $\epsilon$ levels, {\bf\em Texas}, ({\bf\em S4}).}
	\label{fig:SsSA_T}
\end{figure*}

Due to space limitation, in Fig.~\ref{fig:SsSA_T} we plot the multi-attribute subgroup experiments for the Texas dataset, they are discussed in Section~\ref{ssec:SsSA}.

\end{document}